\documentclass{article}
\usepackage{ijcai23}
\usepackage{times}
\usepackage{url}
\usepackage[hidelinks]{hyperref}
\usepackage[utf8]{inputenc}
\usepackage{amsmath}
\usepackage{amsfonts}
\usepackage{amsthm}
\usepackage{booktabs}
\usepackage{algorithm}
\usepackage{algorithmic}
\usepackage{subcaption}
\usepackage{xcolor}
\usepackage{multirow}
\usepackage{graphicx}
\DeclareMathOperator*{\argmax}{arg\,max}

\newtheorem{proposition}{Proposition}

\title{Ensemble Reinforcement Learning in Continuous Spaces -- A Hierarchical Multi-Step Approach for Policy Training}

\author{
Gang Chen$^1$
\and
Victoria Huang$^2$\\
\affiliations
$^1$School of Engineering and Computer Science, Victoria University of Wellington, New Zealand\\ 
$^2$National Institute of Water and Atmospheric Research, New Zealand\\
\emails aaron.chen@ecs.vuw.ac.nz, 
victoria.huang@niwa.co.nz
}

\date{}

\begin{document}

\maketitle

\begin{abstract}
Actor-critic deep reinforcement learning (DRL) algorithms have recently achieved prominent success in tackling various challenging reinforcement learning (RL) problems, particularly complex control tasks with high-dimensional continuous state and action spaces. Nevertheless, existing research showed that actor-critic DRL algorithms often failed to explore their learning environments effectively, resulting in limited learning stability and performance. To address this limitation, several ensemble DRL algorithms have been proposed lately to boost exploration and stabilize the learning process. However, most of existing ensemble algorithms do not explicitly train all base learners towards jointly optimizing the performance of the ensemble. In this paper, we propose a new technique to train an ensemble of base learners based on an innovative multi-step integration method. This training technique enables us to develop a new hierarchical learning algorithm for ensemble DRL that effectively promotes inter-learner collaboration through stable inter-learner parameter sharing. The design of our new algorithm is verified theoretically. The algorithm is also shown empirically to outperform several state-of-the-art DRL algorithms on multiple benchmark RL problems.
\end{abstract}

\section{Introduction}
\label{sec-int}

Deep reinforcement learning (DRL) is a booming field of research in machine learning with diverse real-world applications \cite{ibarz2021}. In recent years, many \emph{model-free} DRL algorithms achieved cutting-edge performance in tackling various continuous reinforcement learning (RL) problems, including complex control tasks with high-dimensional state and action spaces \cite{liu2021}. These algorithms can effectively train \emph{deep neural networks} (DNNs) to precisely model high-quality control policies and are the central focus of this paper\footnote{A long version of this paper with all referenced appendices can be accessed through \url{https://arxiv.org/abs/2209.14488}.}.

Despite of widely reported success, a majority of existing \emph{actor-critic DRL algorithms}, such as DDPG \cite{lillicrap2015}, SAC \cite{haarnoja2018} and PPO \cite{schulman2017}, still suffer from some major limitations. Specifically, existing research works showed that the algorithm performance is highly sensitive to hyper-parameter settings and can vary substantially in different algorithm runs \cite{paine2020}. \emph{Ineffective exploration} is often considered as a major cause for the poor learning stability \cite{chan2019}, often resulting in overfitting and premature convergence to poor local optima \cite{kurutach2018}.

Rather than relying on one learner (or DRL agent), an ensemble of base learners can be jointly utilized to boost exploration and stabilize the learning process \cite{osband2016deep,osband2017post}. For example, the \emph{ensemble deep deterministic policy gradient} (ED2) algorithm is a newly developed ensemble DRL method \cite{januszewski2021} that trains multiple DNN policies simultaneously using a shared \emph{experience replay buffer} (ERB), similar to several previously proposed parallel DRL algorithms \cite{barth2018,mnih2016asyn}. ED2 features a unique mixture of multiple well-studied tricks, including temporally-extended deep exploration, double Q-bias reduction, and target policy smoothing \cite{osband2016deep,osband2017post,van2016deep,fujimoto2018}. It was reported to outperform state-of-the-art ensemble DRL algorithms such as SUNRISE \cite{lee2021sunrise} on several difficult Mujoco benchmark control problems.

As far as we know, many existing ensemble DRL algorithms are designed to train each base learner individually. For example, in ED2, every base learner trains its own DNN policy using its own critic, with the aim to improve its own performance without considering the impact of the trained policy on the ensemble. While sharing the same ERB, policy training is conducted largely independently by all base learners. This is shown to promote healthy exploration in \cite{januszewski2021}. However, there is no guarantee that the base learners will collaborate effectively such that the ensemble as a whole can achieve desirable performance.

To address this limitation, we propose a new \emph{hierarchical approach} for training base learners in this paper. Specifically, we follow ED2 for \emph{low-level training} of DNN policies, which will be performed concurrently by all base learners. In the meantime, we construct a \emph{global critic}, which is trained constantly to predict the performance of the ensemble. Guided by the global critic, \emph{high-level training} of DNN policies will be performed regularly to strengthen cooperation among all the base learners.

Since the ensemble is not used directly to collect state-transition samples from the learning environment, we must make sure that high-level training of the ensemble is not performed on \emph{out-of-distribution} data obtained by individual base learners. In view of this, it is important to encourage \emph{inter-learner parameter sharing} so that the DNN policy trained by one base learner can contribute directly to the training of DNN policies by other base learners. For this purpose, we develop a new technique in this paper for high-level training of policies based on the \emph{multi-step integration methods} \cite{scieur2017}.

Our high-level policy training technique is theoretically justified as it guarantees stability for a wide range of optimization problems. Meanwhile, it can be shown analytically that, for all base learners, their trained linear parametric policies (a special and important technique for policy approximation) are expected to behave more consistently as the ensemble through high-level policy training, encouraging inter-learner collaboration and alleviating the data distribution issue.

Driven by the hierarchical policy training method, we develop a new ensemble DRL algorithm called the \emph{hierarchical ensemble deep deterministic policy gradient} (HED) in this paper. Experimental evaluation of HED has been conducted on a range of benchmark control problems, including the widely used Mujoco control tasks as well as the less popular and potentially more challenging PyBullet control problems. Our experiments clearly show that HED can outperform ED2, SUNRISE and several cutting-edge DRL algorithms on multiple benchmark problems.

\section{Related Work}
\label{sec-rw}

Similar to ED2, HED trains an ensemble of policies using an \emph{off-policy} DRL algorithm to leverage on the algorithm's advantages in \emph{sample efficiency}. Recently, several off-policy DRL algorithms have been developed successfully for RL in continuous spaces, including DDPG \cite{lillicrap2015}, SAC \cite{haarnoja2018}, TD3 \cite{fujimoto2018}, and SOP \cite{wang2020}. These algorithms introduce a variety of tricks to stabilize the learning process. For example, TD3 extends the idea of double Q-network \cite{van2016deep} to a new double-Q bias reduction technique, which can effectively prevent over-optimistic training of DNN policies. In addition, empirical evidence showed that the learning process becomes more stable when the actor and critic in TD3 are trained with different frequencies \cite{fujimoto2018,cobbe2021}. The base learners in our HED ensemble will adopt these tricks.

The recent literature also provides some new tricks to stabilize learning. Specifically, various trust-region methods have been developed to prevent negative behavioral changes during policy training \cite{kurutach2018,schulman2015,shani2020,wu2017,schulman2017}. Meanwhile, entropy regularization techniques prohibit immature convergence of the trained policies and ensure prolonged profitable exploration \cite{chen2018,haarnoja2018}. However, these techniques are mainly applied to stochastic policies while we aim at learning an ensemble of deterministic policies. Previous research showed that deterministic policies can often be trained more efficiently than stochastic policies using the \emph{reparameterization trick} \cite{fujimoto2018,silver2014,baek2020}.

The stability of a DRL algorithm depends critically on how the learner explores its environment. Besides the entropy regularization methods, curiosity metrics are popularly employed to encourage a learner to explore rarely visited states during RL \cite{reizinger2020,zhelo2018}. Meanwhile, many previous studies embraced the \emph{optimum in the face of uncertainty} (OFU) principle to design bonus rewards for actions with high potentials, thereby promoting exploration in promising areas of the learning environment \cite{bellemare2016}. One good example is the UCB exploration technique developed in \cite{chen2017ucb,lee2021sunrise}. However, in \cite{januszewski2021}, this technique was shown to be less effective than the bootstrap with random initialization trick adopted in ED2. Temporally-extended exploration on RL problems with continuous actions can also be achieved by adding a small amount of noise to DNN weights \cite{plappert2017}. This is directly related to the posterior sampling methods that are often used to select the best actions among a statistically plausible set of sampled actions \cite{osband2018}.

Following the OFU principle, deep ensembles have been recently proposed to approximate Bayesian posteriors with high accuracy and efficiency \cite{lakshminarayanan2016}. They are subsequently exploited to approach deep exploration for reliable RL \cite{osband2016deep}. Several issues have been investigated under the context of ensemble DRL. For instance, the diversity of base learners is essential to the performance of the ensemble. To encourage diversity, either different DRL algorithms or the same algorithm with differed hyper-parameter settings have been adopted to train base learners \cite{huang2017,wiering2008}. The training of each base learner can also be supported by an ensemble of critics \cite{an2021uncertainty}. Meanwhile, inter-learner collaboration can be encouraged by asking one learner to imitate the behavior of the other learner that is expected to perform better in the ensemble \cite{lai2020dual}. This idea gives rise to the DPD-PPO algorithm that only supports an ensemble with two learners. Some experiment results that compare the performance of DPD-PPO with HED can be found in Appendix F. 

As far as we know, few existing ensemble DRL algorithms in the literature have ever studied the important issue on how to effectively train all base learners to jointly improve the ensemble performance. This issue will be explored in-depth with the newly developed HED algorithm in this paper.

\section{Background}
\label{sec-back}

An RL problem is modeled as a \emph{Markov Decision Process} (MDP) $(\mathcal{S},\mathcal{A},R,P,\gamma,p_0)$, where $\mathcal{S}$ and $\mathcal{A}$ refer respectively to the continuous multi-dimensional state space and action space. $P$ stands for the state-transition model that governs the probability of reaching any state $s_{t+1}\in\mathcal{S}$ at timestep $t+1$ upon performing any action $a_t\in\mathcal{A}$ in state $s_t\in\mathcal{S}$ at timestep $t$, with $t\in\mathbb{Z}^+$. Additionally, $\gamma\in[0,1)$ is the discount factor, $R$ is the reward function, and $p_0$ captures the initial state distribution.

To solve any RL problem described above, we aim to learn an optimal \emph{deterministic ensemble policy} $\pi^e_*(s)$ that maps any state input $s\in\mathcal{S}$ to an action vector $a\in\mathcal{A}$ so as to maximize the \emph{cumulative rewards} defined below
$$
\pi^e_*= \argmax_{\pi^e} J(\pi^e)=\argmax_{\pi^e}\mathbb{E}_{\tau\sim\pi^e} \left[ \sum_{t=1}^{\infty}\gamma^{t-1} R(s_t,a_t) \right],
$$
where $\tau=[(s_t,a_t,r_t,s_{t+1})]_{t=1}^{\infty}$ contains a series of consecutive state-transition samples and is called a \emph{episode}, which can be obtained by following the ensemble policy $\pi^e$, and $r_t=R(s_t,a_t)$ is the immediate reward received at timestep $t$ in $\tau$. For an ensemble with $N$ base learners where each base learner $L_i$, $1\leq i\leq N$, maintains its own deterministic base policy $\pi^i$, the action output of $\pi^e$ is jointly determined by all the \emph{base policies} according to
\begin{equation}
\forall s\in\mathcal{S}, \pi^e(s)=\frac{1}{N}\sum_i^N \pi^i(s).
\label{equ-pe}
\end{equation}

In order to train an ensemble to maximize the cumulative rewards, our baseline algorithm ED2 uses randomly selected base learners to sample a series of episodes $\{\tau_i\}$, which will be stored in the shared ERB. At regular time intervals, a mini-batch of state-transition samples will be retrieved from the ERB. Every base learner $L_i$ will then use the retrieved mini-batch to train its own actor $\pi^i$ and critic $Q^i$ individually. In other words, a base learner manages two separate DNNs, one models the deterministic policy $\pi^i$ and the other approximates the Q-function $Q^i$ of $\pi^i$. A base learner uses an existing actor-critic RL algorithm to train the two DNNs. In this paper, we choose TD3 for this purpose due to its proven effectiveness, high popularity and stable learning behavior \cite{fujimoto2018}.

\section{Hierarchical Ensemble Deep Deterministic Policy Gradient}
\label{sec-algo}

The pseudo-code of the HED algorithm is presented in Algorithm \ref{alg-code}. HED follows many existing works including ED2 \cite{osband2016deep,januszewski2021} to achieve temporally-extended exploration through bootstrapping with random initialization of DNN policies. As clearly shown in \cite{januszewski2021}, this exploration technique is more effective than UCB and parameter randomization methods. Different from ED2 which completely eliminates the necessity of adding small random noises to the deterministic action outputs from the DNN policies, we keep a small level of action noise\footnote{The noise is sampled from the Normal distribution independently for each dimension of the action vector. The variance of the normal distribution is fixed at 0.01 during the learning process.} while using any chosen policy to explore the learning environment. We found empirically that this ensures coherent exploration, similar to \cite{osband2016deep}, while making the testing performance of the trained policies more stable.

Different from ED2 and other ensemble algorithms for RL in continuous spaces, HED trains DNN policies at two separate levels. The low-level training of $\pi^i$ and $Q^i$ by each base learner $L_i$ is essentially the same as ED2 and TD3. Specifically, for any base learner $L_i$, $i\in\{1,\ldots,N\}$, $Q^i$ is trained by $L_i$ to minimize $MSE_i$ below
\begin{equation}
MSE_i=\frac{1}{|\mathcal{B}|}
\sum_{(s,a,r,s')\in\mathcal{B}}\left(
\begin{array}{l}
Q^i_{\phi_i}(s,a)-r-\\
\displaystyle{\gamma\min_{k=1,2}\hat{Q}^i_{k}(s',\pi^i(s')+\epsilon)}
\end{array}
\right)^2,
\label{equ-mse}
\end{equation}
where $\phi_i$ represents the trainable parameters of the DNN that approximates $Q^i$. $\mathcal{B}$ is the random mini-batch retrieved from the ERB. $\hat{Q}^i_{k}$ with $k=1,2$ stands for the two target Q-networks of $L_i$ that together implement the double-Q bias reduction mechanism proposed in \cite{fujimoto2018}. Additionally, $\epsilon$ is a random noise sampled from a Normal distribution with zero mean and small variance\footnote{The variance for sampling $\epsilon$ is kept at a very small level of 0.01 in the experiments.}. Using the trained $Q^i$, the trainable parameters $\theta_i$ of the DNN that models policy $\pi^i$ is further updated by $L_i$ along the \emph{policy gradient} direction computed below
\begin{equation}
\nabla_{\theta_i}J(\pi^i_{\theta_i})=\frac{1}{|\mathcal{B}|}\sum_{s\in\mathcal{B}} \nabla_{a}Q^i(s,a)|_{a=\pi^i_{\theta_i}(s)}\nabla_{\theta_i}\pi^i_{\theta_i}(s).
\label{equ-pg}
\end{equation}

Besides the above, HED constantly trains a separate high-level Q-function $Q^e$ to predict the performance of the ensemble policy $\pi^e$. Guided by the trained $Q^e$, high-level policy training is conducted regularly to update policy $\pi^i$ of all base learners so as to enhance their cooperation and performance.

A new \emph{multi-step technique} is developed in HED to enable inter-learner parameter sharing during high-level policy training. To implement this technique, we keep track of a list of bootstrap policy parameters for the multi-step training process. More details can be found in the following subsection. Theoretical justifications regarding the usefulness of the multi-step approach are also provided below.

\begin{algorithm}[!ht]
\begin{algorithmic}
\STATE {\bf Input}: Ensemble size $N$; initial policy networks $\pi^i_{\theta_i}$ and Q-networks $Q^i_{\phi_i}$ for $i\in\{1,\ldots,N\}$; ERB; ensemble Q-network $Q^e_{\phi_e}$; target Q-networks for each base learner and the ensemble
\STATE {\bf Output}: Trained ensemble policy $\pi^e$
\STATE {\bf While} total number of sampled trajectories $<$ max number of trajectories:
\STATE \ \ \ \ Randomly sample $i\in\{1,\ldots,N\}$
\STATE \ \ \ \ {\bf While} the current trajectory does not terminate:
\STATE \ \ \ \ \ \ \ \ Use $\pi^i$ to perform the next action
\STATE \ \ \ \ \ \ \ \ Store sampled state-transition in ERB
\STATE \ \ \ \ \ \ \ \ Track number of steps sampled before critic training
\STATE \ \ \ \ \ \ \ \ {\bf If} time for critic training:
\STATE \ \ \ \ \ \ \ \ \ \ \ \ {\bf For} number of steps sampled:
\STATE \ \ \ \ \ \ \ \ \ \ \ \ \ \ \ \ Sample a mini-batch $\mathcal{B}$ from ERB
\STATE \ \ \ \ \ \ \ \ \ \ \ \ \ \ \ \ Train $Q^i_{\phi_i}$ for $i\in\{1,\ldots,N\}$ using \eqref{equ-mse}
\STATE \ \ \ \ \ \ \ \ \ \ \ \ \ \ \ \ Train $Q^e_{\phi_e}$ using \eqref{equ-e-mse}
\STATE \ \ \ \ \ \ \ \ \ \ \ \ \ \ \ \ {\bf If} time for {\bf low-level} policy training:
\STATE \ \ \ \ \ \ \ \ \ \ \ \ \ \ \ \ \ \ \ \ Train $\pi^i_{\theta_i}$ for $i\in\{1,\ldots,N\}$ using \eqref{equ-pg}
\STATE \ \ \ \ \ \ \ \ {\bf If} time for {\bf high-level} policy training:
\STATE \ \ \ \ \ \ \ \ \ \ \ \ Set bootstrap list $\{x_j\}_{j=0}^{2}$ for each base learner
\STATE \ \ \ \ \ \ \ \ \ \ \ \ {\bf For} a fraction of sampled steps:
\STATE \ \ \ \ \ \ \ \ \ \ \ \ \ \ \ \ Train $\pi^i_{\theta_i}$ for $i\in\{1,\ldots,N\}$ using \eqref{equ-mu-new}
\STATE \ \ \ \ \ \ \ \ \ \ \ \ \ \ \ \ Append trained $\theta_i$ for $i\in\{1,\ldots,N\}$ to the
\STATE \ \ \ \ \ \ \ \ \ \ \ \ \ \ \ \ bootstrap lists of each base learner for the next
\STATE \ \ \ \ \ \ \ \ \ \ \ \ \ \ \ \ step of high-level policy training
\end{algorithmic}
\caption{The pseudo-code of the HED algorithm.}
\label{alg-code}
\end{algorithm}

\subsection{Multi-Step High-Level Policy Training}
\label{subsec-multipol}

In addition to $Q^i$ for each base learner $L_i$, $i\in\{1,\ldots,N\}$, HED maintains a separate Q-network to approximate $Q^e$ of the ensemble policy $\pi^e$. Similar to \eqref{equ-mse}, HED trains this central Q-network towards minimizing $MSE_e$ below
\begin{equation}
MSE_e=\frac{1}{|\mathcal{B}|}
\sum_{(s,a,r,s')\in\mathcal{B}}\left(Q^e_{\phi_e}(s,a)-r-\gamma\hat{Q}^e(s',\pi^e(s')) \right)^2,
\label{equ-e-mse}
\end{equation}
with $\phi_e$ representing the trainable parameters of the central Q-network. $\hat{Q}^e$ stands for the corresponding target Q-network that stabilizes the training process. For simplicity, we do not add random noise $\epsilon$ in \eqref{equ-mse} to the action outputs produced by the ensemble policy $\pi^e$ in \eqref{equ-e-mse}. Furthermore, following \cite{van2016deep}, one target Q-network instead of two is adopted in \eqref{equ-e-mse} to facilitate the training of $Q^e$. Building on the trained $Q^e$, we can calculate the \emph{ensemble policy gradient} with respect to $\theta_i$ of every base learner $L_i$ as follows
\begin{equation}
\begin{split}
\nabla_{\theta_i}J(\pi^e)=&
\frac{1}{|\mathcal{B}|}\sum_{s\in\mathcal{B}} \nabla_{a}Q^e(s,a)|_{a=\pi^e(s)}\nabla_{a_i}\pi^e(s)|_{a_i=\pi^i_{\theta_i}(s)}\\
&\nabla_{\theta_i}\pi^i_{\theta_i}(s),
\end{split}
\label{equ-e-pg}
\end{equation}
with
$$
\nabla_{a_i}\pi^e(s)|_{a_i=\pi^i_{\theta_i}(s)}=\frac{1}{N} I,
$$
according to \eqref{equ-pe}. $I$ stands for the $m\times m$ identity matrix where $m$ is the dimension of the action vector. One straightforward approach for high-level policy training is to update $\theta_i$ of every base learner $L_i$ in the direction of \eqref{equ-e-pg}. However, using \eqref{equ-e-pg} alone may not encourage any base learner $L_i$ to behave consistently with the ensemble (see Proposition \ref{the-2}). Consequently, high-level training of the ensemble policy may be performed on the out-of-distribution state-transition samples collected by the base learners, affecting the training effectiveness. Furthermore, ensembles are used mainly for temporally-extended exploration in the literature. Except \cite{lai2020dual}, the learning activity of one base learner may only indirectly influence the learning activities of other base learners through the shared ERB. Base learners do not explicitly share their learned policy parameters to strengthen inter-learner cooperation and boost the learning process.

To address this limitation, we propose to promote inter-learner parameter sharing during high-level policy training, in order to achieve a desirable balance between exploration and inter-learner cooperation. Specifically, in addition to \eqref{equ-e-pg}, we randomly select \emph{two base learners} $L_p$ and $L_q$ and use their policy parameters to guide the training of policy $\pi^i$ of any base learner $L_i$. In comparison to \emph{selecting one base learner}, this allows more base learners to have the opportunity to share their parameters with the base learner $L_i$ during policy training. It is also possible to recruit more than two base learners. However, in this case, it is mathematically challenging to derive stable learning rules for high-level policy training.

Motivated by the above discussion, a search through the literature leads us to the linear multi-step integration methods recently analyzed in \cite{scieur2017}. Consider a simple \emph{gradient flow equation} below
\begin{equation}
x(0)=\theta_i^0, \frac{\partial x(t)}{\partial t}=g(x(t))=\nabla_{\theta_i}J(\pi^e)|_{\theta_i=x(t)},
\label{equ-gfe}
\end{equation}
where $\theta_i^0$ refers to the initial policy parameter of base learner $L_i$. If $J(\pi^e)$ is strongly concave and Lipschitz continuous, the solution of \eqref{equ-gfe} allows us to obtain the optimal policy parameters $\theta_i^*$ when $x(t)$ approaches to $\infty$. Since $J(\pi^e)$ is not strongly concave for most of real-world RL problems, $x(t)$ in practice may only converge to a locally optimal policy, which is common among majority of the policy gradient DRL algorithms. Therefore high-level training of policy $\pi^e$ and hence $\pi^i$ can be approached by numerically solving \eqref{equ-gfe}. This can be achieved through a linear $\mu$-step method shown below
\begin{equation}
x_{k+\mu}=-\sum_{j=0}^{\mu-1}\rho_j x_{k+j}+h\sum_{j=0}^{\mu-1}\sigma_j g(x_{k+j}), \forall k\geq 0,
\label{equ-mu-step}
\end{equation}
where $\rho_j,\sigma_j\in\mathbb{R}$ are the pre-defined coefficients of the multi-step method and $h$ is the learning rate. Clearly, each new point $x_{k+\mu}$ produced by the $\mu$-step method is a function of the preceding $\mu$ points. In this paper, we specifically consider the case when $\mu=3$. Meanwhile, let
\begin{equation}
x_0=\theta_p,x_1=\theta_q,x_2=\theta_i,
\label{equ-boot-list}
\end{equation}
where $p$ and $q$ are the randomly generated indices of two base learners and $i$ is the index of the base learner whose policy $\pi^i$ is being trained by the $\mu$-step method. Through this way, the training of policy $\pi^i$ is influenced directly by base learners $L_p$ and $L_q$ through explicit inter-learner parameter sharing. $x_i (i\geq 3)$ in \eqref{equ-mu-step} represents the trained policy parameters of $\pi^i$ in subsequent training steps.

Although \eqref{equ-mu-step} allows us to use $\nabla_{\theta_p}J(\pi^e)$ and $\nabla_{\theta_q}J(\pi^e)$ to train $\theta_i$, they do not seem necessary for inter-learner parameter sharing. To simplify \eqref{equ-mu-step}, we set $\sigma_0=\sigma_1=0$ and $\sigma_2=1$. Hence only $g(x_{k+2})$, which is the ensemble policy gradient with respect to policy $\pi^i$ in \eqref{equ-e-pg}, is used to train $\pi^i$. With this simplification, we derive the new learning rule for high-level policy training below
\begin{equation}
\begin{split}
& x_{k+3}= -\rho_2 x_{k+2}-\rho_1 x_{k+1}-\rho_0 x_k+h\cdot \nabla_{\theta_i}J(\pi^e)|_{\theta_i=x_{k+2}} \\
& x_0=\theta_p,x_1=\theta_q,x_2=\theta_i, \forall k\geq 0.
\end{split}
\label{equ-mu-new}
\end{equation}

To implement \eqref{equ-mu-new} in HED, before high-level policy training, every base learner $L_i$ must set up a \emph{bootstrap list} of policy parameters $\{x_0=\theta_p,x_1=\theta_q,x_2=\theta_i\}$. After the $k$-th training step ($k\geq 0$) based on \eqref{equ-mu-new}, $L_i$ appends the trained $\theta_i$ as $x_{k+3}$ to the bootstrap list, which will be utilized to train $\pi^i$ in the subsequent training steps. Reliable use of \eqref{equ-mu-new} demands for careful parameter settings of $\rho_0$, $\rho_1$, $\rho_2$ and $h$. Relevant theoretical analysis is presented below.

\subsection{Theoretical Analysis}
\label{subsec-multithe}

In this subsection, a theoretical analysis is performed first to determine suitable settings of $\rho_0$, $\rho_1$, $\rho_2$ and $h$ for stable high-level policy training through \eqref{equ-mu-new}. To make the analysis feasible, besides the strongly concave and Lipschitz continuous conditions, we further assume that
\begin{equation}
\nabla_{\theta_i}J(\pi^e)\approx -A(\theta_i-\theta_i^*)
\label{equ-grad-linear}
\end{equation}
where $A$ is a positive definite matrix whose eigenvalues are bounded positive real numbers. $\theta_i^*$ stands for the global-optimal (or local-optimal) policy parameters. Many strongly concave functions satisfy this assumption \cite{scieur2017}. Meanwhile, the attraction basin of the local optimum of many multi-modal optimization problems often satisfies this assumption too. Using this assumption, we can derive Proposition \ref{the-1} below.

\begin{proposition}
Upon using \eqref{equ-mu-new} to numerically solve \eqref{equ-gfe}, the following conditions must be satisfied for $x_k$ to converge to $\theta_i^*$ as $k$ approaches to $\infty$:
\begin{enumerate}
    \item $\rho_2=\rho_0-1$, $\rho_1=-2\rho_0$;
    \item $0<\rho_0<\frac{1}{2}$;
    \item $h$ is reasonably small such that $0\leq \lambda h < 2-4\rho_0$, where $\lambda$ can take any real value between the smallest and the largest eigenvalues of the positive definite matrix $A$ in \eqref{equ-grad-linear}.
\end{enumerate}
\label{the-1}
\end{proposition}

The proof of Proposition \ref{the-1} can be found in Appendix A. Proposition \ref{the-1} provides suitable parameter settings for \eqref{equ-mu-new} and justifies its stable use for high-level policy training. We next show that \eqref{equ-mu-new} is also expected to make base learners behave more consistently with the ensemble, without affecting the behavior of the trained ensemble, when $\rho_0$ is sufficiently small. Consider specifically that each base learner $L_i$ trains a linear parametric policy of the form:
\begin{equation}
\pi^i(s)=\Phi(s)^T\cdot \theta_i
\label{equ-lin-pol}
\end{equation}
where $\Phi(s)$ represents the \emph{state feature vector} with respect to any input state $s$. For simplicity, we study the special case of scalar actions. However, the analysis can be easily extended to high-dimensional action spaces. Meanwhile, we use $Sin()$ and $Mul()$ to represent respectively the action output of a policy trained for one iteration on the same state $s$ by using either the single-step method or the multi-step method in \eqref{equ-mu-new}. The single-step method can be considered as a special case of the multi-step method with $\rho_2=-1$ and $\rho_0=\rho_1=0$. Using these notations, Proposition \ref{the-2} is presented below.

\begin{proposition}
When each base learner $L_i$, $i\in\{1,\ldots,N\}$, trains its linear parametric policy $\pi^i$ with policy parameters $\theta_i$ on any state $s\in\mathcal{S}$ and when $0<\rho_0<\frac{1}{3}$,
\begin{enumerate}
    \item $Sin(\pi^e(s))=\mathbb{E}\left[ Mul(\pi^e(s)) \right]$;
    \item $\sum_{i\in\{1,\ldots,N\}}\mathbb{E}\left[ \left( Mul(\pi^i(s))-\mathbb{E}\left[ Mul(\pi^e(s)) \right] \right)^2 \right] \\ < \sum_{i\in\{1,\ldots,N\}}\left( Sin(\pi^i(s))-Sin(\pi^e(s)) \right)^2 \\ =\sum_{i\in\{1,\ldots,N\}}\left( \pi^i(s)-\pi^e(s) \right)^2$
\end{enumerate}
where the expectations above are taken with respect to any randomly selected $p,q\in\{1,\ldots,N\}$ in \eqref{equ-boot-list}.
\label{the-2}
\end{proposition}

Proposition \ref{the-2} indicates that multi-step training in \eqref{equ-mu-new} is expected to reduce the difference between the action output of any base learner and that of the ensemble. Meanwhile the amount of action changes applied to $\pi^e$ remains identical to the single-step method. Therefore, using the multi-step policy training method developed in this section helps to enhance consistent behaviors among all base learners of the ensemble.

\begin{table*}[!ht]
\caption{Final performance of all competing algorithms on 9 benchmark problems. The results are shown with mean cumulative rewards and standard deviation over 10 independent algorithm runs. For each run, the cumulative rewards are obtained by averaging over 50 independent testing episodes.}
\centering
\resizebox{1.0\textwidth}{!}{
\begin{tabular}{c||cccc|c}
\hline
Benchmark problems                       & TD3            & SAC            & ED2             & SUNRISE         & HED             \\ \hline 
Ant-v0 (PyBullet)              & 3246.82$\pm$184.03  & 2453.23$\pm$523.96  &  3285.06$\pm$183.98  &  2425.93$\pm$1120.12  &  \textbf{3370.12$\pm$179.95}\\
Hopper-v0 (PyBullet)              & 2051.68$\pm$567.12  & 2126.43$\pm$165.58  & 2284.41$\pm$203.79  & 1585.17$\pm$761.78  & \textbf{2530.85$\pm$277.26}   \\
InvertedPendulum-v0 (PyBullet)       & 958.15$\pm$33.38   & 995.69$\pm$12.94     & \textbf{1000.00$\pm$0.00}      & 995.69$\pm$5.27     & \textbf{1000.00$\pm$0.00}      \\
Walker2D-v0 (PyBullet)              & 1379.56$\pm$394.77  & 774.19$\pm$281.45    &  1082.42$\pm$312.4 &  2012.15$\pm$148.73 & \textbf{2109.49$\pm$116.79}\\
Hopper-v3 (Mujoco)             & 2374.47$\pm$721.68 & 3306.01$\pm$410.28 & 2361.13$\pm$1101.71  & 2427.12$\pm$622.42  & \textbf{3396.9$\pm$249.11}   \\
Humanoid-v3   (Mujoco)            & 321.1$\pm$225.3  &  1151.49$\pm$598.54 & 596.57$\pm$113.03 &  756.66$\pm$143.06  & \textbf{4223.77$\pm$1119.55}  \\
InvertedDoublePendulum-v2 (Mujoco)               & 7417.5$\pm$3694.8   & \textbf{9355.67$\pm$11.04}    & 9323.89$\pm$22.15   & 9351.58$\pm$22.15      & 9144.38$\pm$327.28   \\
LunarLanderContinuous-v2         & 275.69$\pm$5.99  &  277.67$\pm$3.44 &  275.7$\pm$7.2  &  282.84$\pm$1.99  & \textbf{286.29$\pm$0.8} \\
Walker2D-v3 (Mujoco)             & 3805.88$\pm$1402.27 & 4240.35$\pm$694.64 &  4750.82$\pm$530.3 &  5510.83$\pm$669.98 &  \textbf{5778.84$\pm$133.15}  \\ \hline
\end{tabular}}
\label{tab:final_perf_comp}
\end{table*}

\begin{figure*}[!hbt]
    \begin{center}
        \subfloat[Ant-v0 (PyBullet)]{\label{fig-hc}\includegraphics[width=0.33\linewidth]{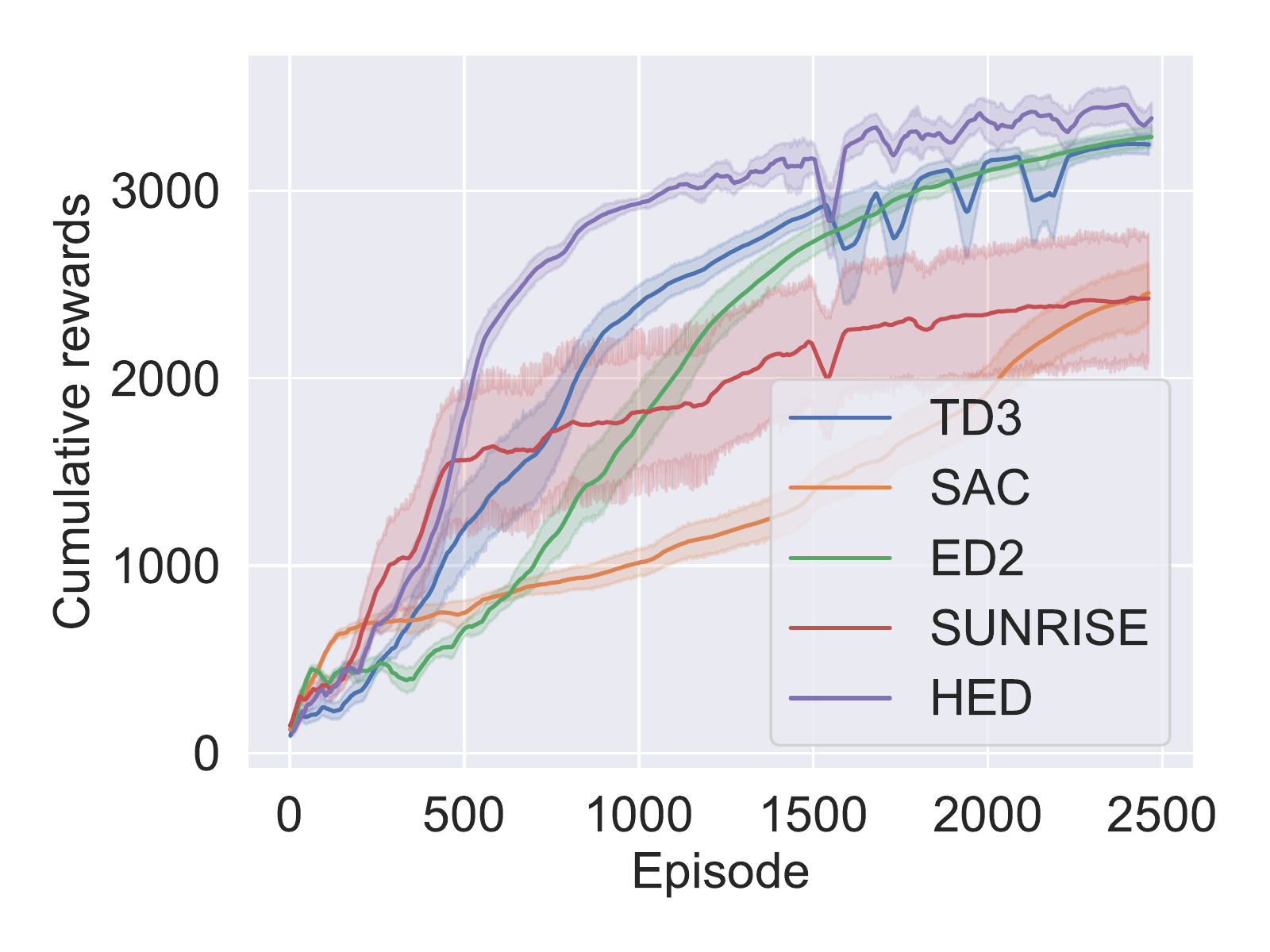}}
        \subfloat[Hopper-v0 (PyBullet)]{\label{fig-hc}\includegraphics[width=0.33\linewidth]{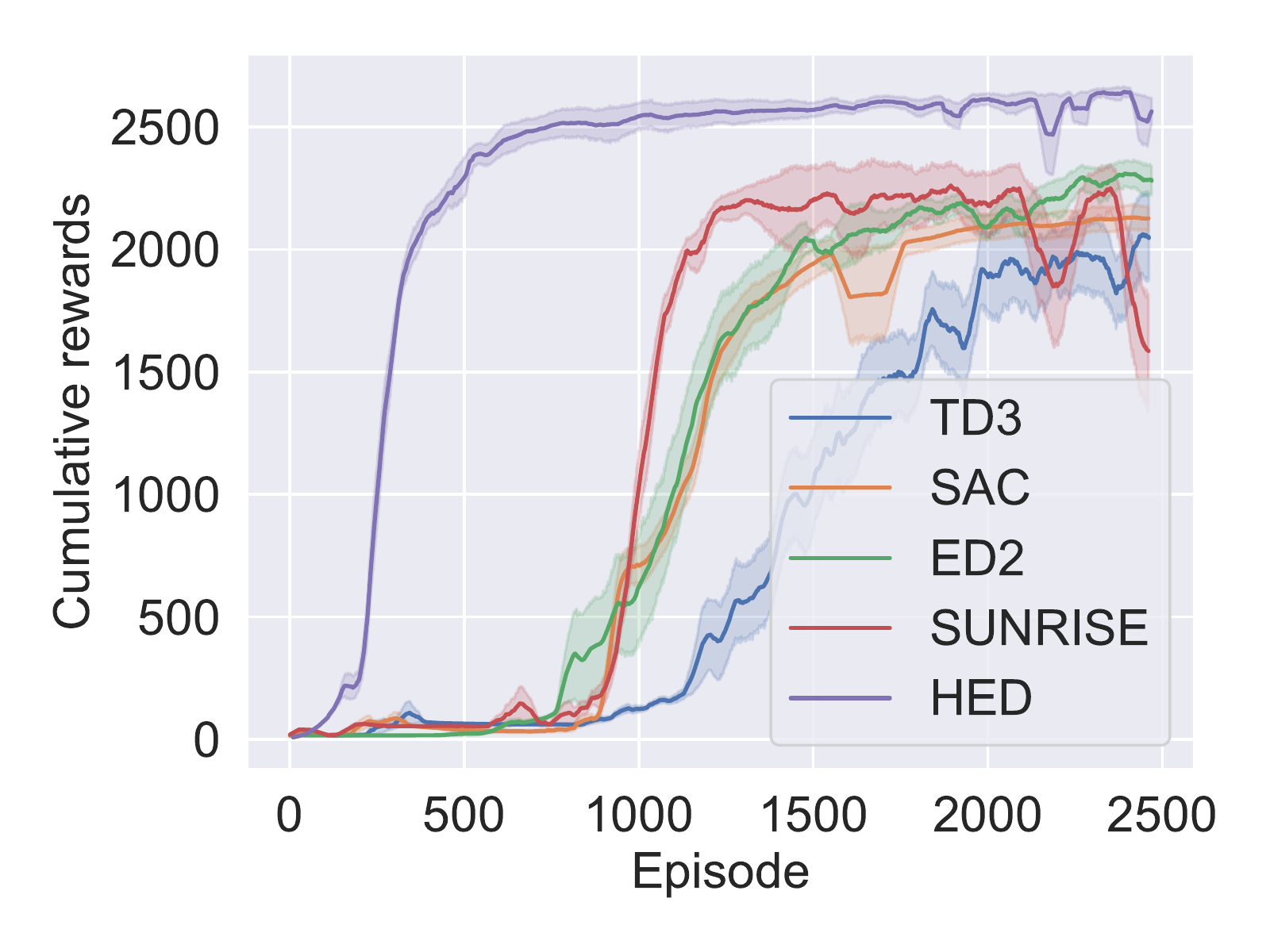}}
       \subfloat[InvertedPendulum-v0 (PyBullet)]{\label{fig-ipPB}\includegraphics[width=0.33\linewidth]{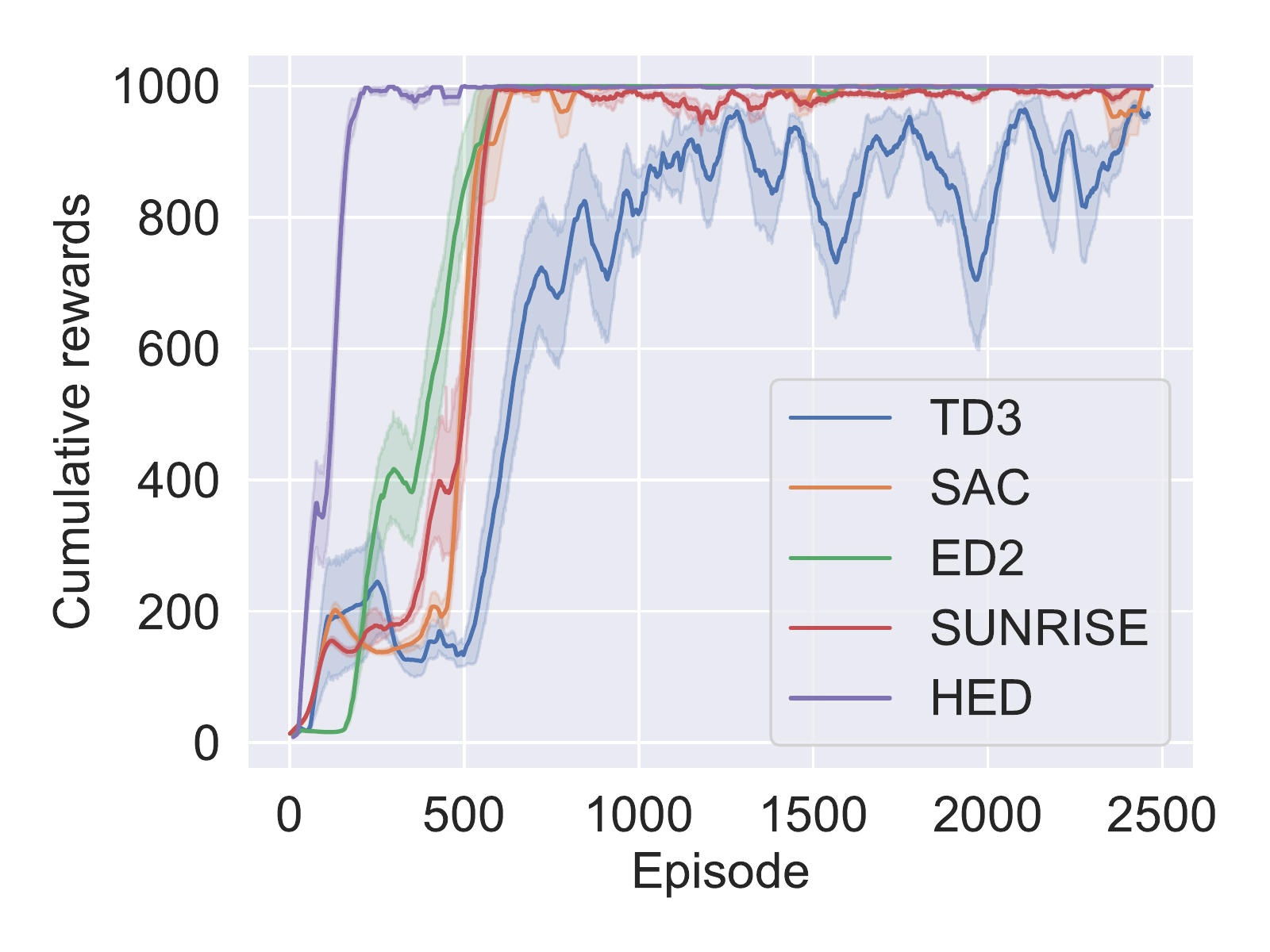}}\\
       \subfloat[Walker2D-v0 (PyBullet)]{\label{fig-reacher}\includegraphics[width=0.33\linewidth]{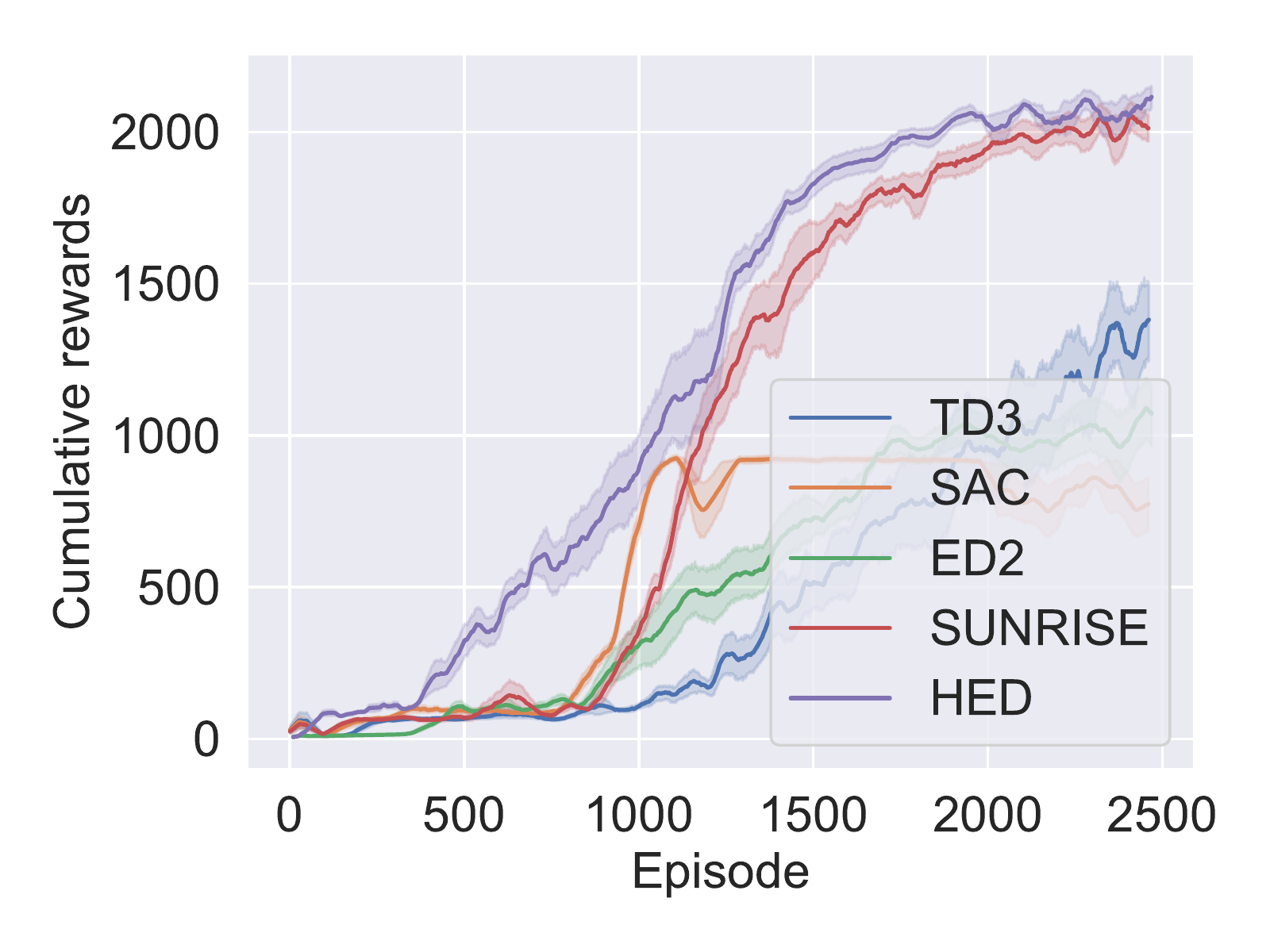}}
       \subfloat[Hopper-v3 (Mujoco)]{\label{fig-hopper}\includegraphics[width=0.33\linewidth]{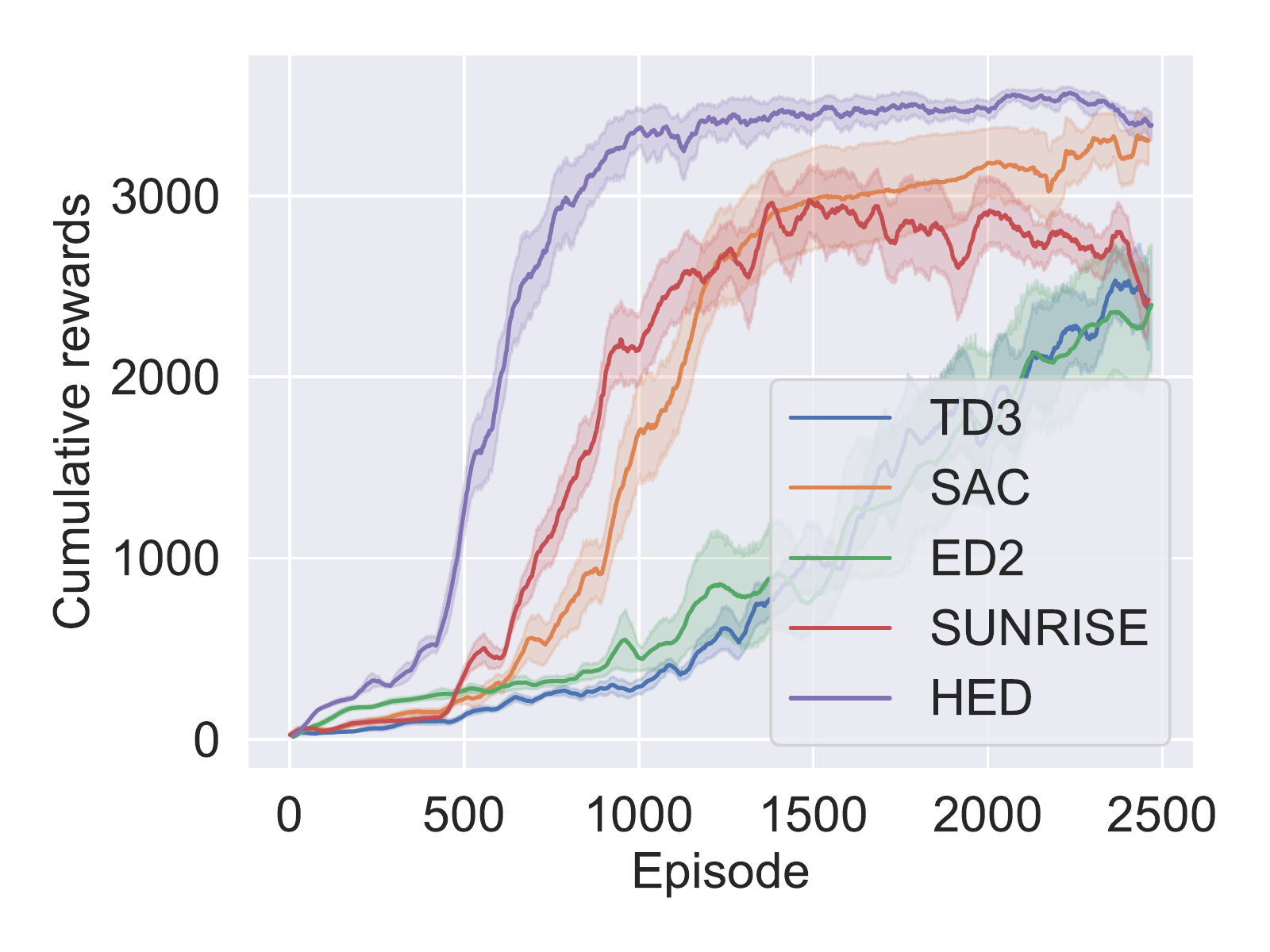}}
       \subfloat[Humanoid-v3 (Mujoco)]{\label{fig-humanoid}\includegraphics[width=0.33\linewidth]{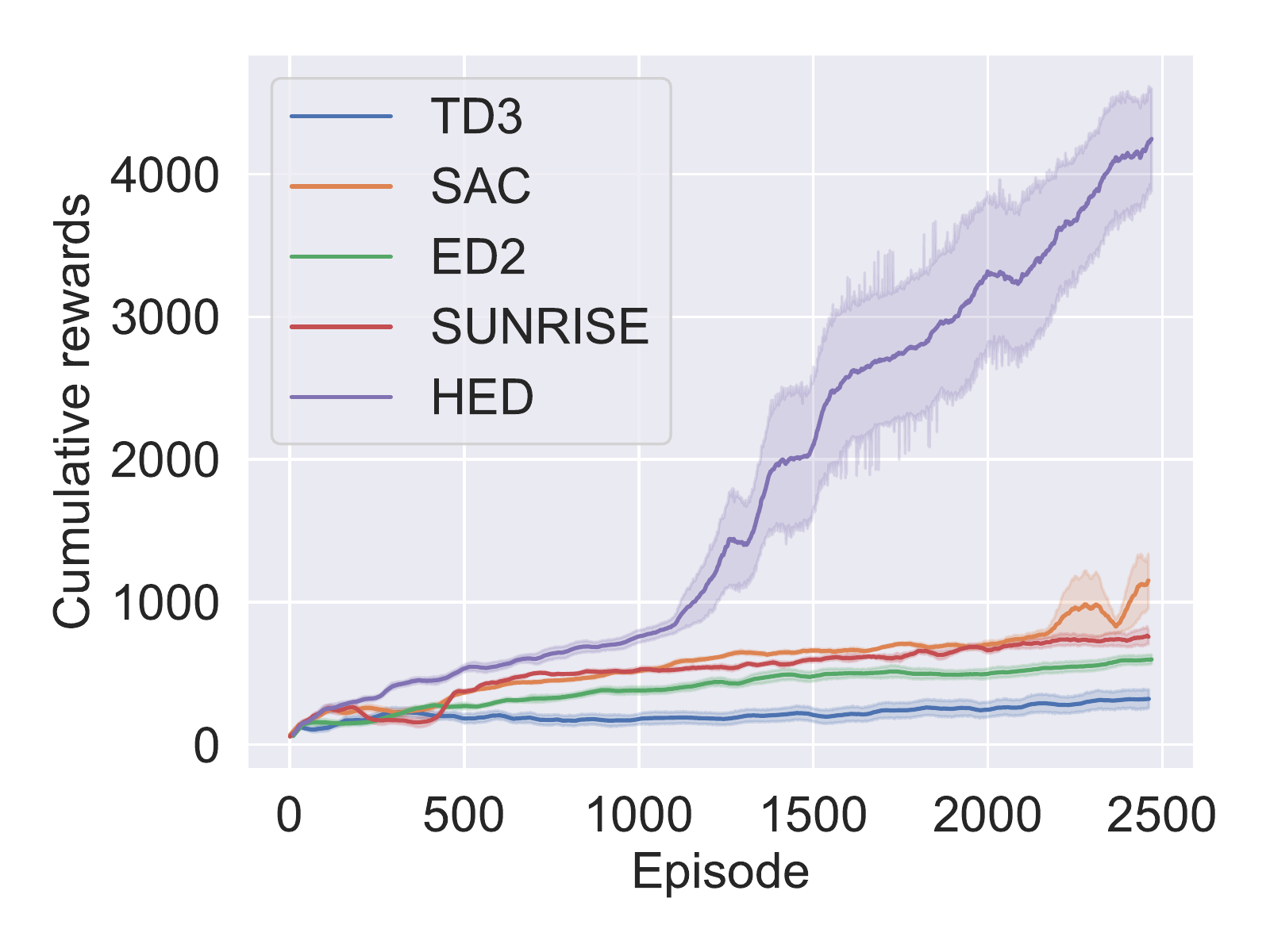}}\\
       \subfloat[InvertedDoublePendulum-v2 (Mujoco)]{\label{fig-humanoid}\includegraphics[width=0.33\linewidth]{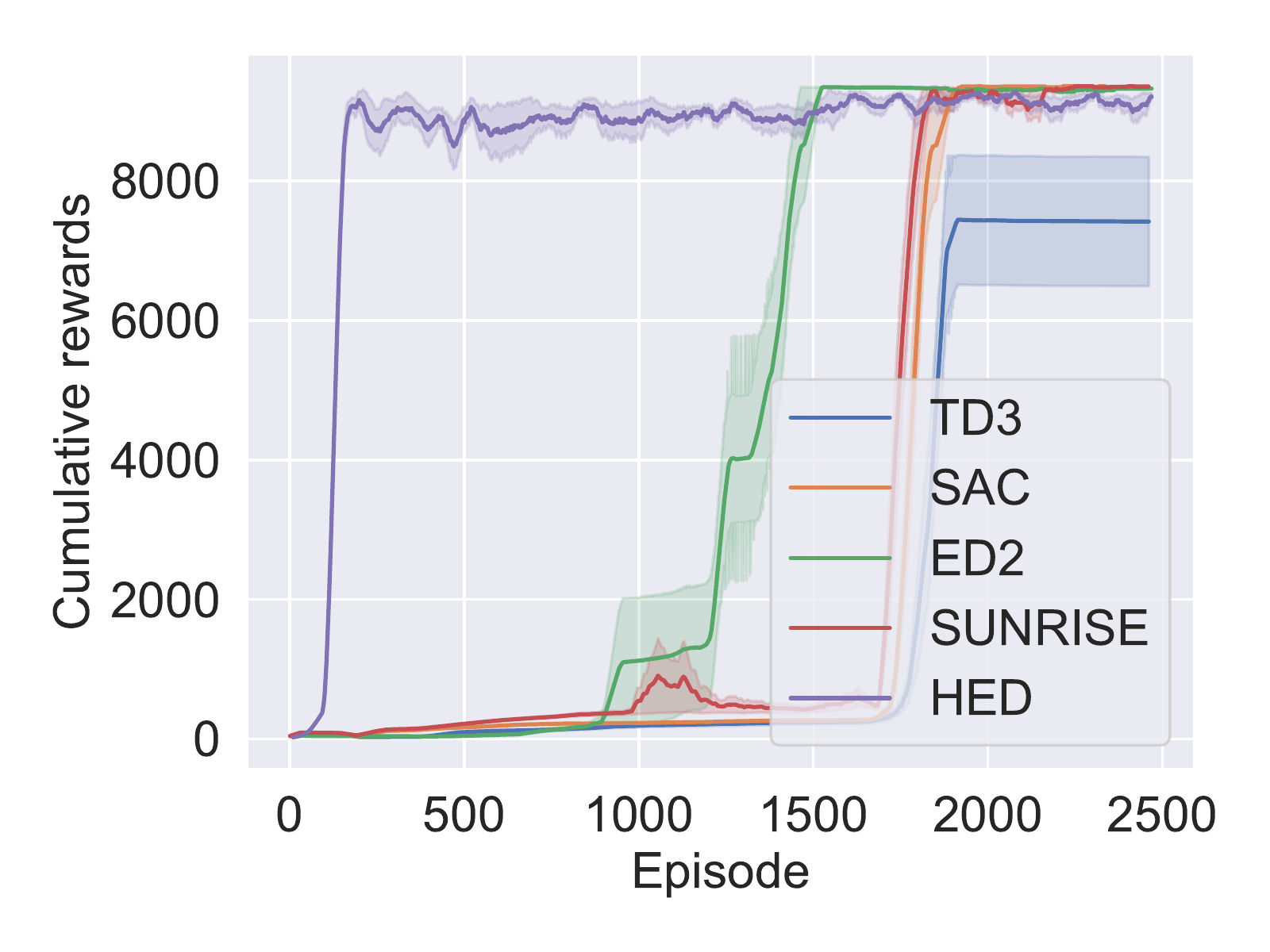}}
       \subfloat[LunarLanderContinuous-v2]{\label{fig-lunarLander}\includegraphics[width=0.33\linewidth]{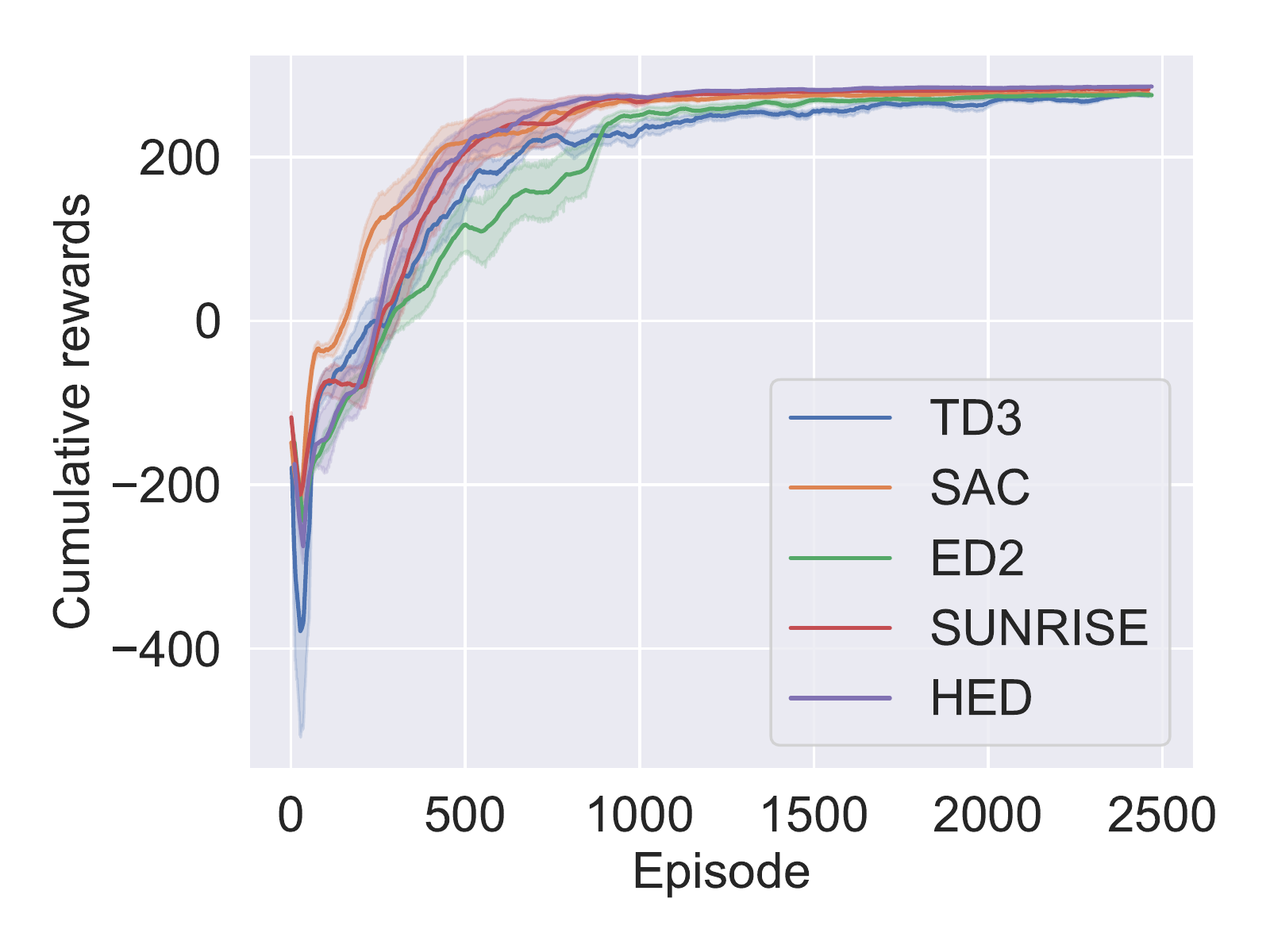}}
       \subfloat[Walker2D-v3 (Mujoco)]{\label{fig-walker}\includegraphics[width=0.33\linewidth]{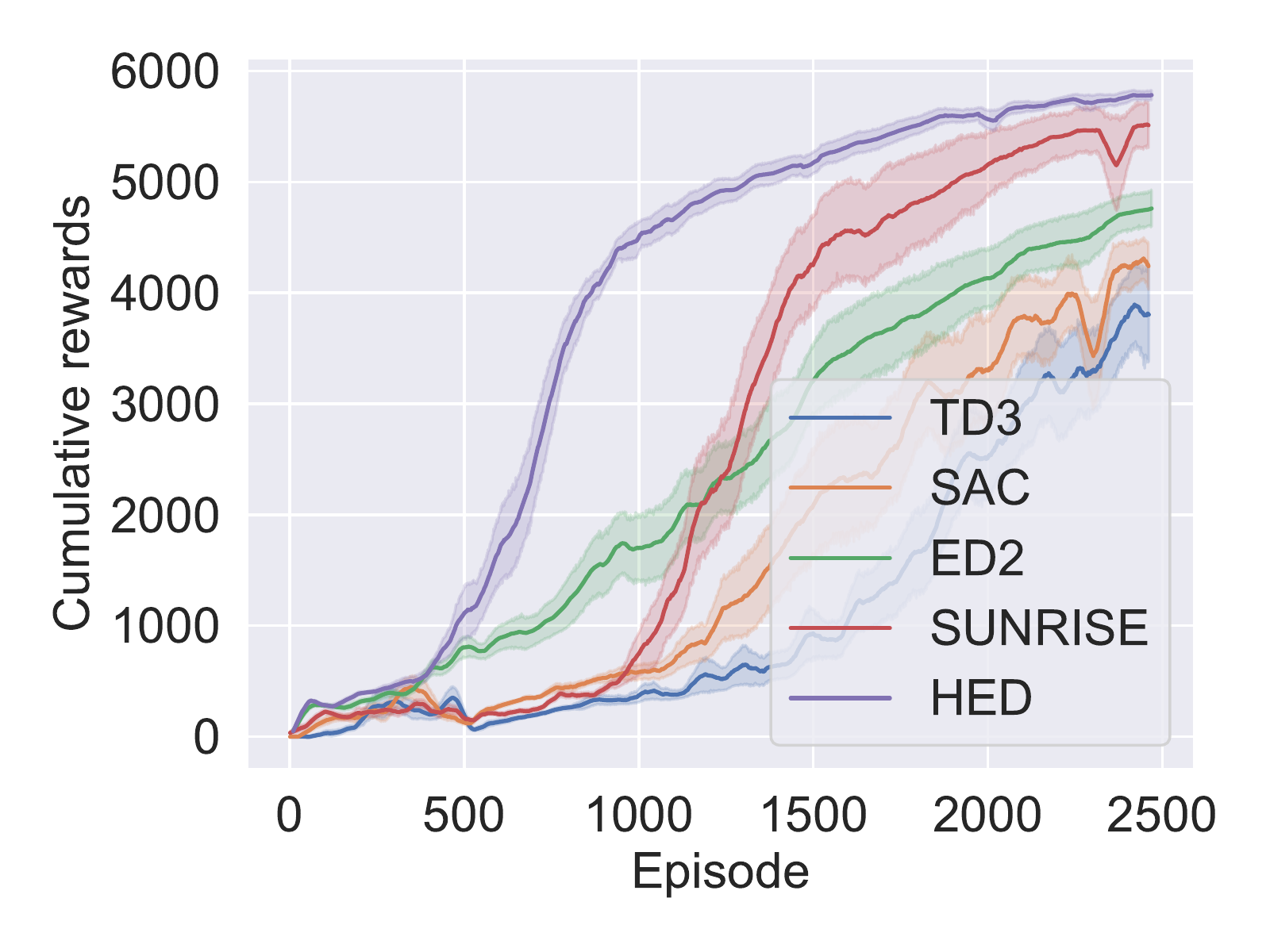}}
       \\
    \end{center}
    \caption{Learning curves of HED and four baseline algorithms (i.e., TD3, SAC, ED2 and SUNRISE) on 9 benchmark RL problems.}
    \label{fig:training_perf}
\end{figure*}


\section{Experiment}
\label{sec-exp}

This section presents the experimental evaluation of HED, in comparison to several state-of-the-art DRL algorithms. The experiment setup is discussed first. Detailed experiment results are further presented and analyzed.

\subsection{Experiment Setting}

We implement HED based on the high-quality implementation of TD3 provided by the publicly available OpenAI Spinning Up repository~\cite{SpinningUp2018}. We also follow closely the hyper-parameter settings of TD3 recommended in \cite{fujimoto2018} to build each base learner of HED. Specifically, a fully connected MLP with two hidden layers of 256 ReLU units is adopted to model all policy networks and Q-networks. Similar to \cite{januszewski2021,lee2021sunrise}, HED employs $5$ base learners, i.e., $N=5$. Each base learner has its own policy network and Q-network. Meanwhile, HED maintains and trains a separate ensemble Q-network with the same network architecture design.

Each base learner trains its Q-network and also conducts the low-level training of the policy network repeatedly whenever HED collects 50 consecutive state-transition samples from the learning environment. Meanwhile, high-level policy training as well as the training of the ensemble Q-network is performed immediately after HED samples a full episode. HED adopts a separate Adam optimizer with the fixed learning rate of $1\mathrm{e}{-3}$ to train each Q-network and policy network. Furthermore, $\rho_0$ in \eqref{equ-mu-new} is set to 0.0001 for the main experiment results reported in Figure \ref{fig:training_perf}. The mini-batch size $|\mathcal{B}|$ is set to 256, following existing research \cite{januszewski2021,fujimoto2018} without any fine-tuning.

HED is compared against four state-of-the-art DRL algorithms, including two Ensemble DRL algorithms, i.e., ED2~\cite{januszewski2021} and SUNRISE~\cite{lee2021sunrise}), and two widely used off-policy DRL algorithms, i.e., SAC~\cite{haarnoja2018} and TD3~\cite{fujimoto2018}. We evaluate their performance on 9 challenging continuous control benchmark problems, including four PyBullet benchmark problems~\cite{benelot2018} (i.e., Ant-v0, Hopper-v0, InvertedPendulum-v0, and Walker2D-v0), five Mujoco control tasks (i.e., Hopper-v3, Humanoid-v3, InvertedDoublePendulum-v0, and Walker2D-v3), and LunarLanderContinuous-v2 provided by OpenAI Gym~\cite{openai_gym}. In literature, PyBullet benchmarks are often considered to be more challenging than Mujoco benchmarks. Hence we decide to evaluate the performance of HED on both PyBullet and Mujoco benchmarks. The maximum episode length for each benchmark is fixed to 1000 timesteps. Each algorithm runs independently with 10 random seeds on all benchmarks. Besides the hyper-parameter settings of HED highlighted above, more detailed hyper-parameter settings of all competing algorithms have been summarized in Appendix~C. 

\subsection{Experiment Result}

\subsubsection{Performance Comparison}
Table~\ref{tab:final_perf_comp} presents the average cumulative rewards obtained by the policy networks (or policy ensembles for ensemble DRL algorithms) trained by all the competing algorithms across the same number of sampled episodes with respect to each benchmark. As evidenced in the table, HED achieved consistently the best performance\footnote{HED significantly outperformed ED2 on most benchmark problems, thanks to its use of the proposed high-level policy training technique.} on most of the benchmark problems except InvertedDoublePendulum.
Meanwhile, on InvertedDoublePendulum, HED achieved very competitive performance with at least 97\% of the highest cumulative rewards reached by the best performing competing algorithms. Furthermore, on some problems such as Humanoid-v3, HED outperformed the lowest performing algorithm by up to 1200\% and the algorithm with the second highest performance by up to 600\%. Besides the results on the average cumulative rewards, the maximum cumulative rewards achieved by each algorithm have been reported in Appendix F for all experimented benchmarks.

In addition to Table~\ref{tab:final_perf_comp}, we also compared the learning curves of all the competing algorithms in Figure \ref{fig:training_perf}. As demonstrated in this figure, by explicitly strengthening inter-learner collaboration, HED converges clearly faster and is more stable during the learning process than other competing algorithms. Specifically, on several benchmark problems, such as Hopper-v0, InvertedPendulum-v0, Hopper-v3, InvertedDoublePendulum, and Walker2D-v3, HED achieved significantly higher sample efficiency and lower variations in learning performance across 10 independent runs. In comparison to other ensemble DRL algorithms, the learning curves of HED also appear to be smoother on several benchmark problems, such as Hopper-v0 and Walker2D-v3, suggesting that HED can achieve highly competitive stability during learning.

\subsubsection{Performance Impact of $\rho_0$}
\begin{figure}
    \begin{center}
        \subfloat[Ant-v0]{\label{fig-ant-rho}\includegraphics[width=0.5\linewidth]{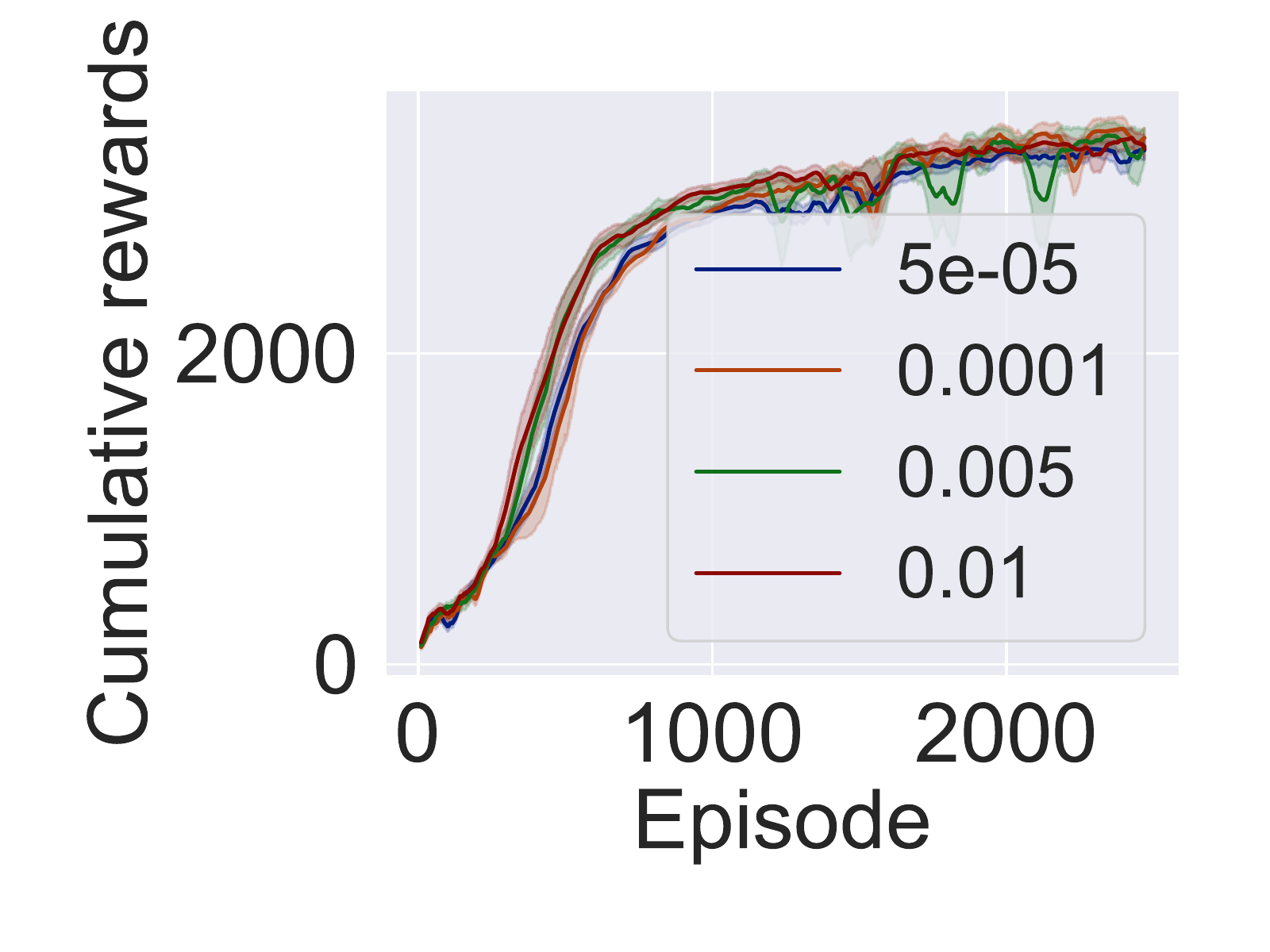}}
        \subfloat[Hopper-v0]{\label{fig-hp-rho}\includegraphics[width=0.5\linewidth]{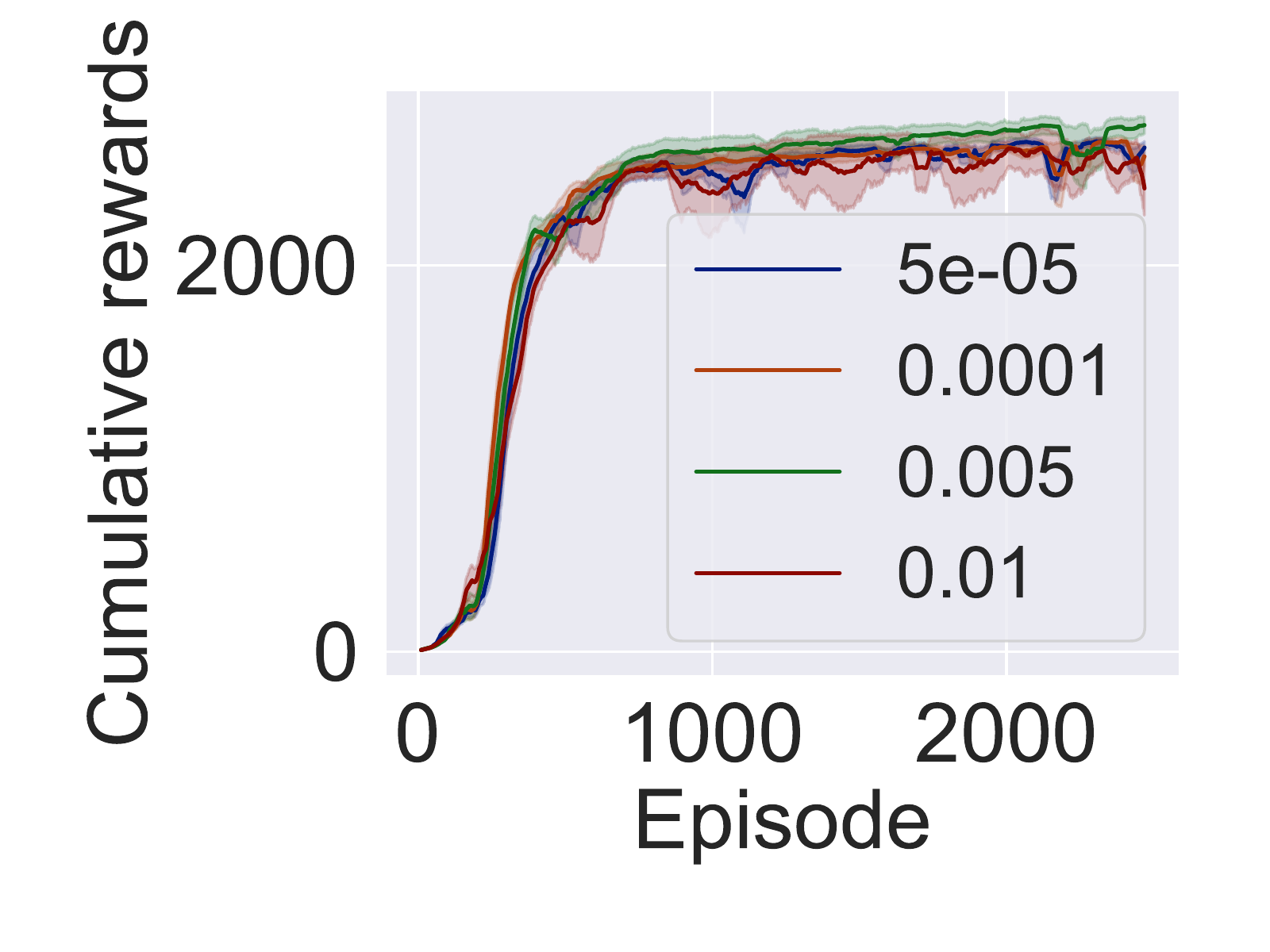}}
    \end{center}
    \caption{The impact of using different $\rho_0$ in \eqref{equ-mu-new} on the performance of HED. $\rho_1$ and $\rho_2$ in \eqref{equ-mu-new} depend directly on $\rho_0$ according to Proposition \ref{the-1}.}
    \label{fig:rho_impact}
\end{figure}

To investigate the performance impact of $\rho_0$, we tested 4 different settings of $\rho_0$, ranging from $5\mbox{e$-$}05$ to $0.01$, on the Ant-v0 and Hopper-v0 problems (similar observations can be found on other benchmark problems and are omitted in this paper). The learning curves are plotted in Figure~\ref{fig:rho_impact}. It is witnessed in the figure that
the impact of different $\rho_0$ on the final performance appears to be small as long as $\rho_0$ is reasonably small according to Proposition \ref{the-1}.

\subsubsection{Ablation Study on High-Level Policy Training Techniques}

High-level policy training can be conducted repeatedly whenever HED obtains either a full sampled episode or a fixed number of consecutive state-transition samples (e.g., samples collected from 50 consecutive timesteps). To understand which approach is more effective, experimental comparisons have been conducted in Appendix D with detailed performance results. According to the experiment results in Appendix D, episodic learning can produce more stable learning behavior and also makes HED converge faster with higher performance.

We also compared HED with its variation that performs high-level policy training by using the single-step method in \eqref{equ-e-pg} instead of the multi-step method in \eqref{equ-mu-new}. Detailed experiment results can be found in Appendix E. Our experiment results confirm that multi-step training in \eqref{equ-mu-new} enables HED to achieve significantly faster convergence and learning stability than using the conventional single-step training technique in \eqref{equ-e-pg}. Hence, by explicitly sharing learned policy parameters among base learners in an ensemble through \eqref{equ-mu-new}, HED can effectively enhance inter-learner collaboration and boost the learning process.

\section{Conclusions}
\label{sec-con}

In this paper, we conducted in-depth study of ensemble DRL algorithms, which have achieved cutting-edge performance on many benchmark RL problems in the recent literature. Different from existing research works that rely mainly on each base learner of an ensemble to train its policy network individually, we developed a new HED algorithm to explore the potential of training all base learners in a hierarchical manner in order to promote inter-learner collaboration and improve the collective performance of an ensemble of trained base learners. Specifically, we adopted existing ensemble DRL algorithms such as ED2 to perform low-level policy training. Meanwhile, a new multi-step training technique was developed for high-level policy training in HED to facilitate direct inter-learner parameter sharing. Both theoretical and empirical analysis showed that the HED algorithm can achieve stable learning behavior. It also outperformed several state-of-the-art DRL algorithms on multiple benchmark RL problems.



\bibliographystyle{named}
\bibliography{ijcai22}

\newpage
\appendix

\section{Proof of Proposition \ref{the-1}}

This appendix presents a proof of Proposition \ref{the-1}. According to \cite{scieur2017}, any multi-step integration methods including \eqref{equ-mu-new} must satisfy three conditions to ensure its stability. They together guarantee that $x_k$ can converge to $\theta_i^*$ as $k$ approaches to $\infty$. We check each condition one-by-one below to derive the main conclusions in Proposition \ref{the-1}.

\vspace{0.2cm}
\noindent
\textbf{Consistency condition}: We can re-write \eqref{equ-mu-new} as below
$$
x_{k+3}+\rho_2 x_{k+2}+\rho_1 x_{k+1}+\rho_0 x_k= h\cdot g(x_{k+2}).
$$
Define the \emph{shift operator} $F$, which maps $Fx_k\rightarrow x_{k+1}$. Furthermore, with $g(x_k)$ being simplified to $g_k$, $F$ also maps $Fg_k\rightarrow g_{k+1}$. Using $F$, \eqref{equ-mu-new} can be further written as
$$
\rho(F)x_k=h\sigma(F)g_k, \forall k\geq 0,
$$
where
$$
\rho(F)=F^3+\rho_2 F^2+\rho_1 F+\rho_0, \sigma(F)=F^2.
$$

The consistency condition requires
$$
\rho(1)=0, \rho'(1)=\sigma(1).
$$
This implies that
\begin{equation*}
\begin{split}
& 1+\rho_2+\rho_1+\rho_0=0, \\
& 3+2\rho_2+\rho_1=1.
\end{split}
\end{equation*}
Solving the above equations leads to
$$
\rho_1=-2\rho_0,\ \rho_2=\rho_0-1.
$$
Hence, \eqref{equ-mu-new} becomes
$$
x_{k+3}=(1-\rho_0)x_{k+2}+2\rho_0 x_{k+1} -\rho_0 x_k+h\cdot g(x_{k+2}).
$$

\vspace{0.2cm}
\noindent
\textbf{Zero-stability condition}: This condition requires all roots of $\rho(F)$ to be in the unit disk. Any roots on the unit circle must be simple. In other words,
$$
\left| Roots(\rho(F)) \right|\leq 1.
$$
In fact, $\rho(F)$ has three roots. They are
$$
1, \frac{1}{2}\left(-\rho_0 \pm \sqrt{\rho_0(\rho_0+4)} \right).
$$
It is easy to verify that when $0<\rho_0<\frac{1}{2}$,
$$
\left| \frac{1}{2}\left(-\rho_0 - \sqrt{\rho_0(\rho_0+4)} \right) \right|<1.
$$
Meanwhile, when $\rho_0>0$,
$$
\left| \frac{1}{2}\left(-\rho_0 + \sqrt{\rho_0(\rho_0+4)} \right) \right|<1.
$$
In summary, the zero-stability condition requires
$$
0<\rho_0<\frac{1}{2}.
$$

\vspace{0.2cm}
\noindent
\textbf{Absolute stability condition}: Define
\begin{equation*}
\begin{split}
\Pi_{\lambda h} & \overset{\Delta}{=}\rho(F)+\lambda h\sigma(F) \\
&=F^3+\rho_2 F^2+\rho_1 F+\rho_0+\lambda h F^2 \\
&=F^3+(\rho_0-1) F^2 - 2\rho_0 F+\rho_0 +\lambda h F^2 \\
&=F^3 + (\rho_0-1+\lambda h)F^2-2\rho_0 F+\rho_0.
\end{split}
\end{equation*}
Further define
$$
r_{max}=\max_{\lambda\in [L,U]} \max_{r\in Roots(\Pi_{\lambda h}(F))}|r|,
$$
where $L$ and $U$ in this appendix refer respectively to the smallest and the largest positive eigenvalues of matrix $A$ in \eqref{equ-grad-linear}. The absolute stability condition requires
\begin{equation}
r_{max}<1.
\label{equ-abs-con}
\end{equation}
Let
\begin{equation*}
\begin{split}
& B=\rho_0-1+\lambda h, \\
& C=-2\rho_0, \\
& D=\rho_0.
\end{split}
\end{equation*}
Subsequently, define
\begin{equation*}
\begin{split}
& A_0=1-B+C-D=2-\lambda h-4\rho_0, \\
& A_1=3-B-C-3D=4-\lambda h -2\rho_0, \\
& A_2=3+B-C-3D=2+\lambda h, \\
& A_3=1+B+C+D=\lambda h.
\end{split}
\end{equation*}
According to the Routh-Hurwitz criterion \cite{nise2020}, the following two conditions jointly guarantee \eqref{equ-abs-con}:
\begin{equation*}
\begin{split}
A_1,A_2,A_3,A_4>0, \\
A_1 A_2 > A_0 A_3.
\end{split}
\end{equation*}
Specifically, the first condition above gives rise to the following:
$$
\lambda h >0,\ \lambda h+2\rho_0<4,\ \lambda h+4\rho_0<2.
$$
Following the second condition above, we can deduce the below:
$$
\lambda h>2-\frac{4}{\rho_0}.
$$
Given that $\lambda h>0$, we have
\begin{equation*}
\begin{split}
& \lambda h > \max\left\{ 0, 2-\frac{4}{\rho_0} \right\}, \\
& \lambda h < \min\left\{ 2-4\rho_0, 4-2\rho_0 \right\}.
\end{split}
\end{equation*}
Since $0<\rho_0<\frac{1}{2}$,
$$
2-4\rho_0<4-2\rho_0,\ 2-\frac{4}{\rho_0}<0.
$$
Consequently
$$
0<\lambda h<2-4\rho_0.
$$
Clearly, with sufficiently small $h$, the above condition on absolute stability can be easily satisfied. Hence, we can use  \eqref{equ-mu-new} to perform high-level policy training stably in the HED algorithm.

\section{Proof of Proposition \ref{the-2}}

This appendix presents a proof of Proposition \ref{the-2}. Considering any specific state $s\in\mathcal{S}$, let
$$
\nabla_a Q^e(s,a)|_{a=\pi^e(s)}=C,
$$
where $C$ is an arbitrary scalar constant, in line with the assumption of scalar actions. Using \eqref{equ-pe} and \eqref{equ-lin-pol}, the ensemble policy gradient with respect to policy parameters $\theta_i$ of policy $\pi^i$, $i\in[1,\ldots,N]$, is
$$
\nabla_{\theta_i} J(\pi^e)=\frac{C\Phi(s)}{N}.
$$
According to the multi-step learning rule in \eqref{equ-mu-new}, updating $\theta_i$ for one iteration gives the updated $\theta_i$ as
$$
(1-\rho_0)\theta_i+2\rho_0\theta_q-\rho_0\theta_p+h\frac{C\Phi(s)}{N}.
$$
Therefore,
\begin{equation*}
\begin{split}
Mul(\pi^i(s))=&(1-\rho_0)\pi^i(s)+2\rho_0\pi^q(s)-\rho_0\pi^p(s) \\
&+\frac{hC}{N}\Phi(s)^T\Phi(s).
\end{split}
\end{equation*}
Hence
$$
\mathbb{E}\left[ Mul(\pi^i(s)) \right]=(1-\rho_0)\pi^i(s)+\rho_0\pi^e(s)+\frac{h C}{N}\Phi(s)^T\Phi(s),
$$
$$
\mathbb{E}\left[ Mul(\pi^e(s)) \right]=\pi^e(s)+\frac{h C}{N}\Phi(s)^T\Phi(s).
$$
In comparison, upon using the single-step method, the updated $\theta_i$ becomes
$$
\theta_i+h\frac{C\Phi(s)}{N}.
$$
Subsequently,
$$
Sin(\pi^i(s))=\pi^i(s)+\frac{h C}{N}\Phi(s)^T\Phi(s),
$$
$$
Sin(\pi^e(s))=\pi^e(s)+\frac{h C}{N}\Phi(s)^T\Phi(s).
$$
Clearly,
$$
Sin(\pi^e(s))=\mathbb{E}\left[ Mul(\pi^e(s)) \right].
$$
Hence, the expected action changes applied to $\pi^e(s)$ are identical, regardless of whether single-step or multi-step method is used for high-level policy training\footnote{We assume in Proposition \ref{the-2} that high-level policy training is performed for one iteration on a specific state $s$.}.

Define
$$
\Delta=\sum_{i\in[1,\ldots,N]}(\pi^i(s)-\pi^e(s)).
$$
For the single-step method, after all base learners trained their respective policies for one iteration on state $s$, it is easy to verify that
\begin{equation*}
\begin{split}
& \sum_{i\in[1,\ldots,N]}\left( Sin(\pi^i(s))-Sin(\pi^e(s)) \right)^2\\
=& \sum_{i\in[1,\ldots,N]}\left( \pi^i(s)-\pi^e(s) \right)^2\\
=& \Delta.
\end{split}
\end{equation*}
Meanwhile,
\begin{equation*}
\begin{split}
& \left( Mul(\pi^i(s))-\mathbb{E}\left[ Mul(\pi^e(s)) \right] \right)^2 \\
=&(
(1-\rho_0) (\pi^i(s)-\pi^e(s)) + 2\rho_0 (\pi^q(s)-\pi^e(s))
\\
&-\rho_0(\pi^p(s)-\pi^e(s)))^2\\
=&(1-\rho_0)^2(\pi^i(s)-\pi^e(s))^2+4\rho_0^2(\pi^q(s)-\pi^e(s))^2\\
&+\rho^2(\pi^p(s)-\pi^e(s))^2\\
&+4(1-\rho_0)\rho_0(\pi^i-\pi^e(s))(\pi^q(s)-\pi^e(s)) \\
&-2(1-\rho_0)\rho_0(\pi^i(s)-\pi^e(s))(\pi^p(s)-\pi^e(s))\\
&-4\rho_0^2(\pi^q(s)-\pi^e(s))(\pi^p(s)-\pi^e(s)).
\end{split}
\end{equation*}
Since the base learner indices $p$ and $q$ are randomly and independently selected,
$$
\mathbb{E}\left[(\pi^i(s)-\pi^e(s))(\pi^p(s)-\pi^e(s))\right]=0,
$$
$$
\mathbb{E}\left[(\pi^i(s)-\pi^e(s))(\pi^q(s)-\pi^e(s))\right]=0,
$$
$$
\mathbb{E}\left[(\pi^q(s)-\pi^e(s))(\pi^p(s)-\pi^e(s))\right]=0.
$$
Therefore
\begin{equation*}
\begin{split}
& \sum_{i\in[1,\ldots,N]} \mathbb{E}\left[
\left( Mul(\pi^i(s))-\mathbb{E}\left[ Mul(\pi^e(s)) \right] \right)^2
\right] \\
=&(1-\rho_0)^2\Delta+4\rho_0^2\Delta+\rho_0^2\Delta \\
=&(1-2\rho_0+6\rho_0^2)\Delta.
\end{split}
\end{equation*}
When $0<\rho_0<\frac{1}{3}$,
$$
1-2\rho_0+6\rho_0^2<1.
$$
As a result,
\begin{equation*}
\begin{split}
&\sum_{i\in[1,\ldots,N]}\mathbb{E}\left[ \left( Mul(\pi^i(s))-\mathbb{E}\left[ Mul(\pi^e(s)) \right] \right)^2 \right] \\
<&\sum_{i\in[1,\ldots,N]}\left( Sin(\pi^i(s))-Sin(\pi^e(s)) \right)^2.
\end{split}
\end{equation*}

\section{Hyper-Parameter Setting}

Table~\ref{tab:hyper-para} provides detailed hyper-parameter settings of all algorithms. Our hyper-parameter settings follow strictly the recommended settings in \cite{fujimoto2018,haarnoja2018,januszewski2021,lee2021sunrise}.

\begin{table}[htb!]
\caption{Hyper-parameter settings of all experimented algorithms.}
\label{tab:hyper-para}
\centering
\resizebox{1.05\linewidth}{!}{
\begin{tabular}{l||lllll}
\hline
Hyper-parameter                    & TD3              & SAC              & ED2   & SUNRISE & {\bf HED} \\ \hline
Num. episodes                        & 2500             & 2500             & 2500  & 2500 & 2500    \\
Episode length                         & 1000             & 1000             & 1000  & 1000   & 1000 \\
Minibatch size                     & 256              & 256              & 256   & 256  &  256 \\
Adam learning rate                 & 3e-4             & 3e-4             & 1e-4  & 3e-4  &  1e-3 \\
Discount ($\gamma$)                       & 0.99             & 0.99             & 0.99  & 0.99  & 0.99  \\
GAE parameter ($\lambda$)                  & 0.995            & 0.995            & 0.995 & 0.995  & 0.995 \\
Replay buffer size                 & 1e6              & 1e6              & 1e6   & 1e6     & 1e6 \\
Update interval                    & 50               & 50               & 50    & 50    & 50    \\
Ensemble size                      & - & - & 5     & 5  & 5 \\ 
Network architecture & 256x256  & 256x256 & 256x256 & 256x256 & 256x256 \\
\hline
\end{tabular}}
\end{table}

\begin{figure*}[!hbt]
    \begin{center}
        \subfloat[Ant-v0 (PyBullet)]{\label{fig-antPB-when2HLtrain}\includegraphics[width=0.33\linewidth]{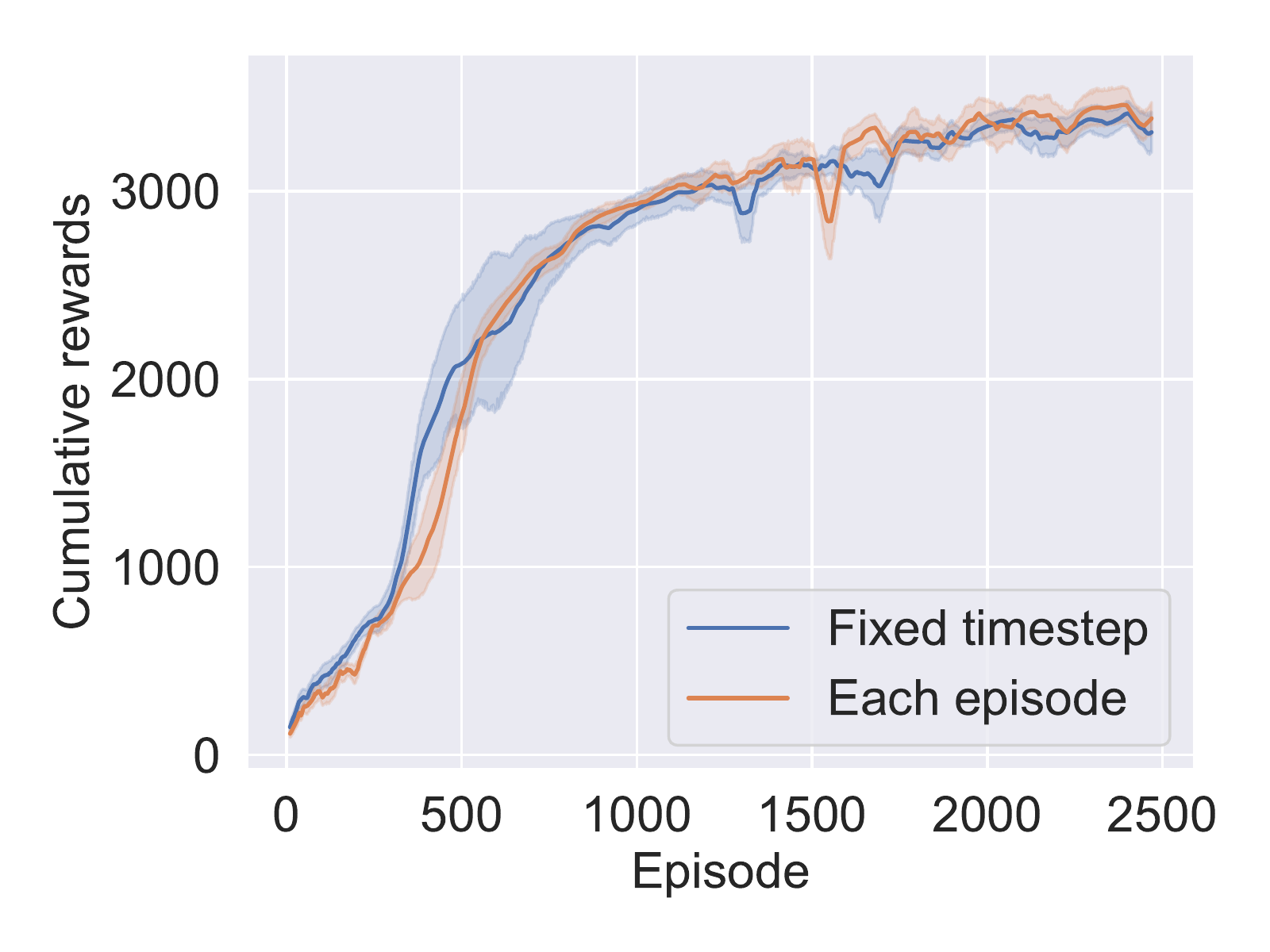}}
        \subfloat[Hopper-v0 (PyBullet)]{\label{fig-hc-when2HLtrain}\includegraphics[width=0.33\linewidth]{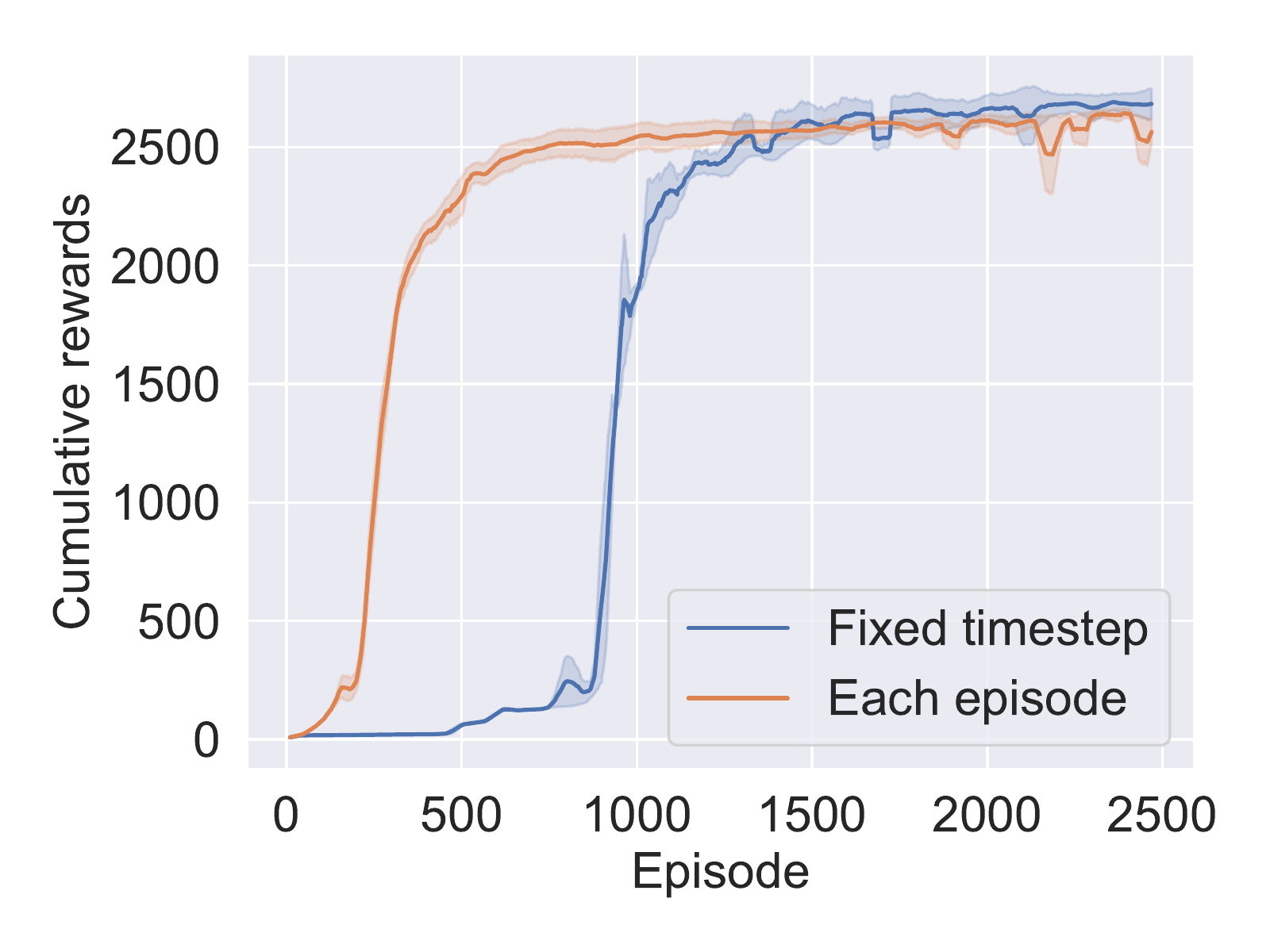}}
        \subfloat[Walker2D-v0 (PyBullet)]{\label{fig-walkerPB-when2HLtrain}\includegraphics[width=0.33\linewidth]{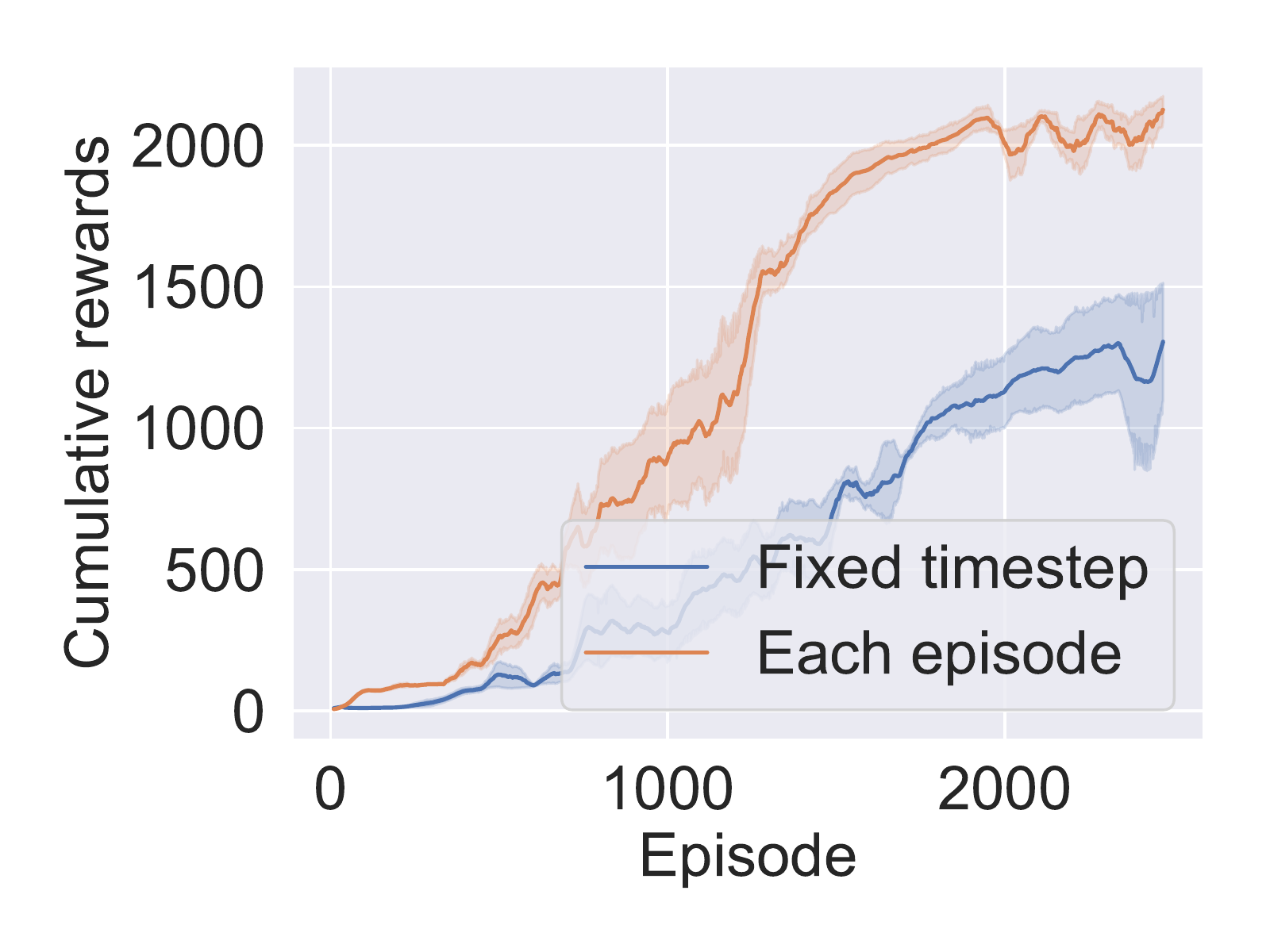}} \\
      \subfloat[InvertedDoublePendulum-v2 (Mujoco)]{\label{fig-idpPB-when2HLtrain}\includegraphics[width=0.33\linewidth]{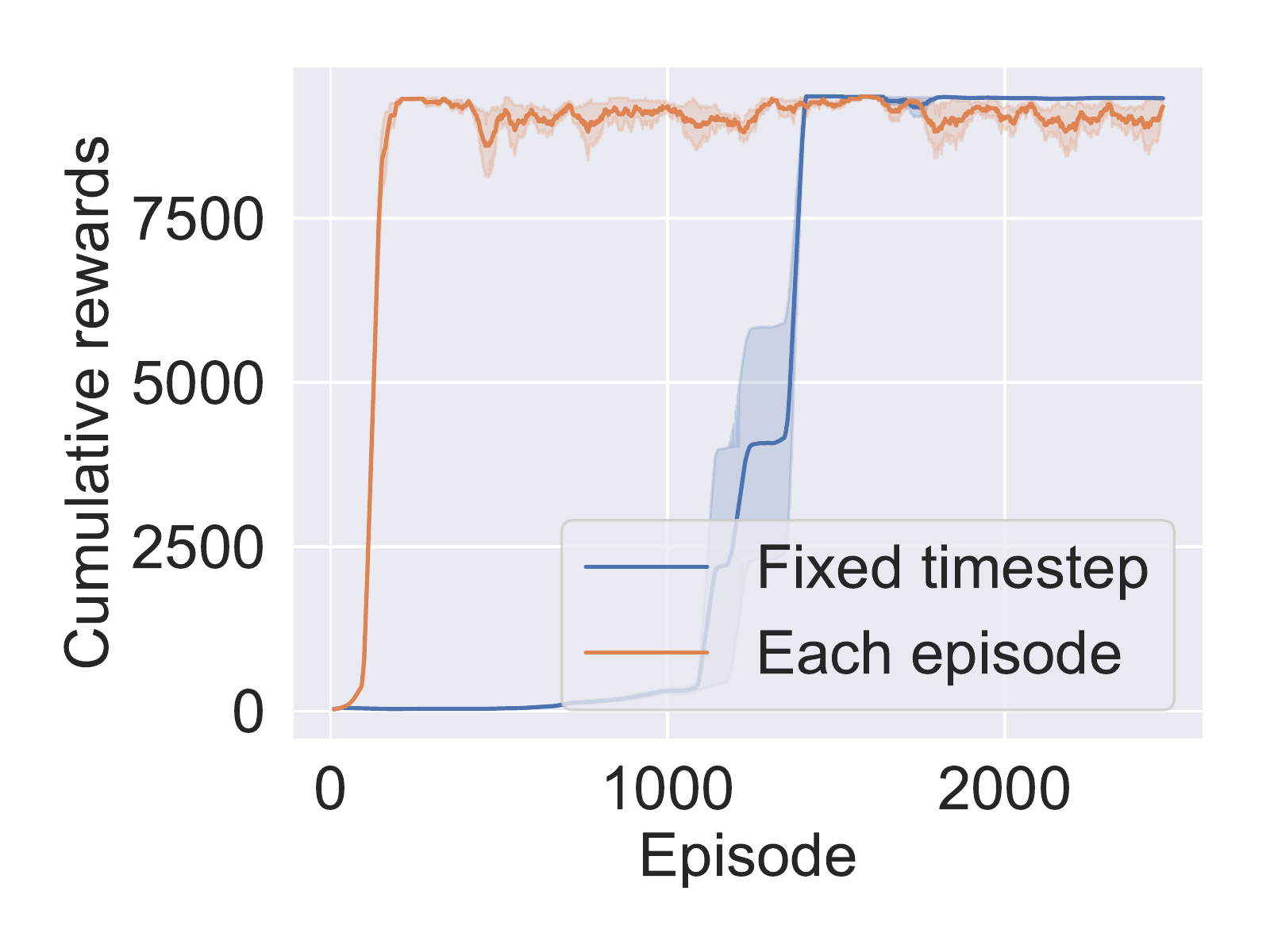}}
       \subfloat[Hopper-v3 (Mujoco)]{\label{fig-hopper-when2HLtrain}\includegraphics[width=0.33\linewidth]{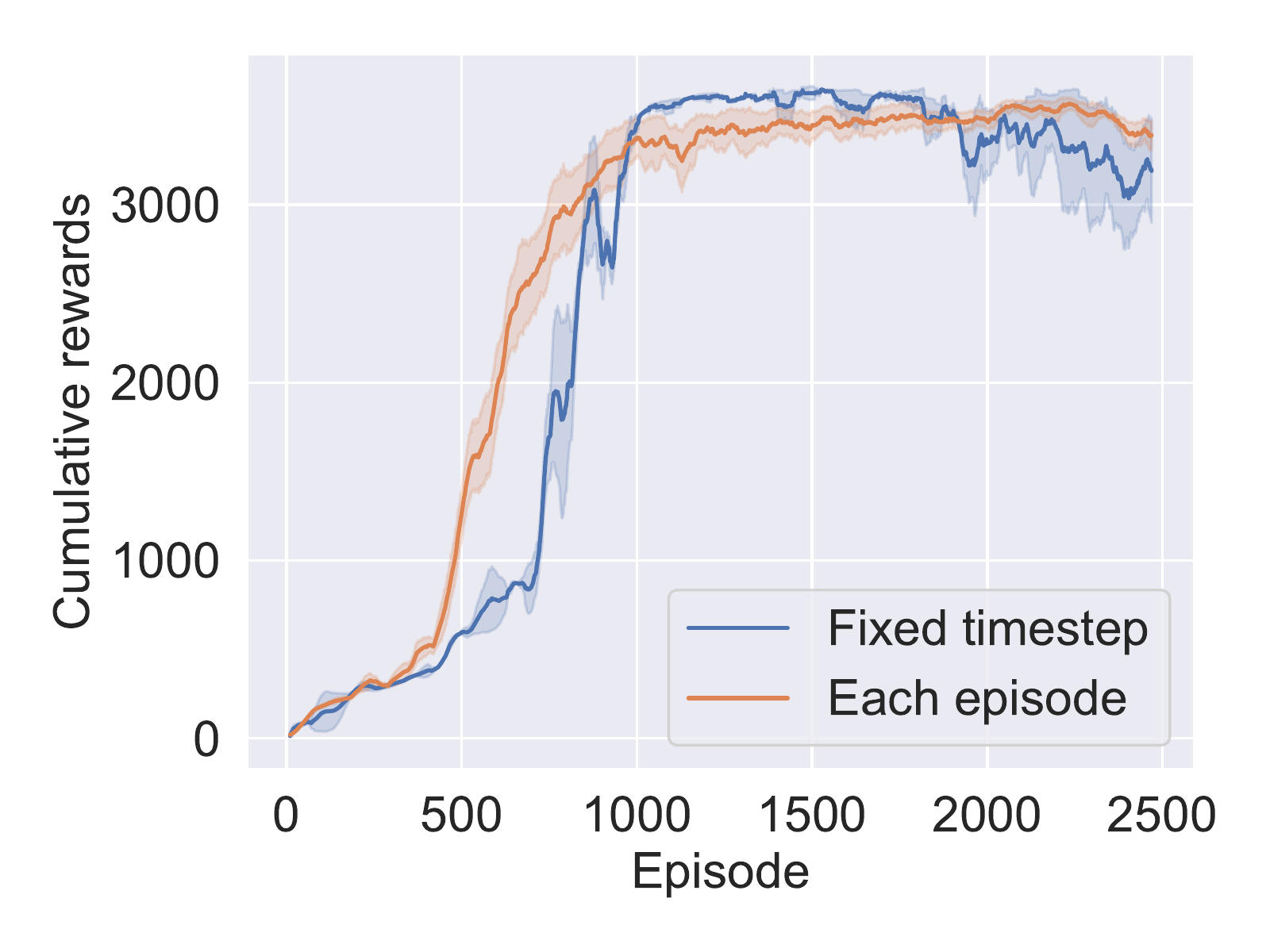}}
       \subfloat[Walker2D-v3 (Mujoco)]{\label{fig-walkerPB-when2HLtrain}\includegraphics[width=0.33\linewidth]{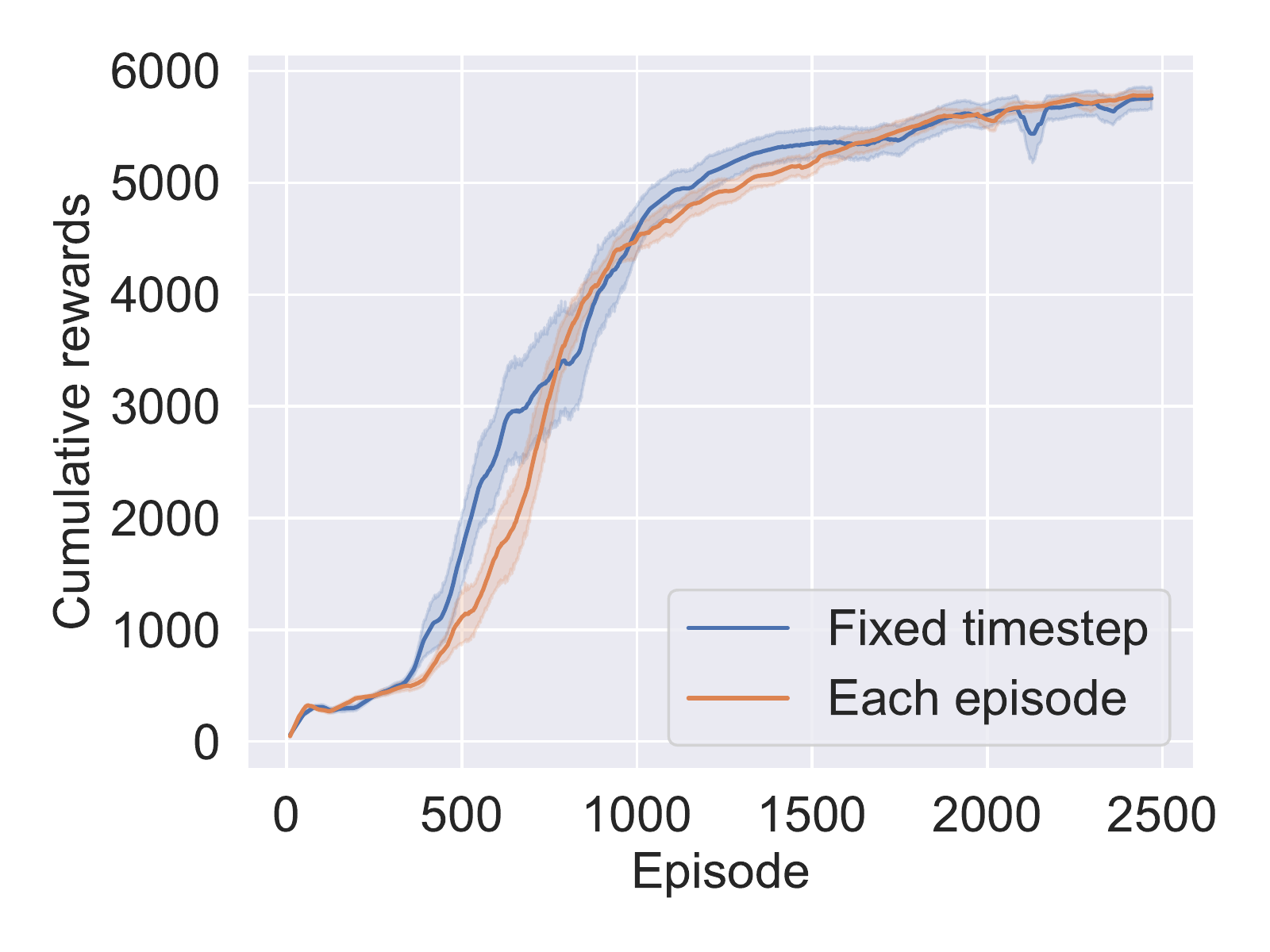}}
\end{center}
    \caption{Learning curves of HED with respect to two high-level policy training approaches. The method that conducts high-level policy training after every 50 timesteps is denoted as ``\emph{Fixed timestep}". The method that conducts high-level policy training at the end of each sampled episode is denoted as ``\emph{Each episode}".}
    \label{fig:when2HLtrain_impact}
\end{figure*}

\begin{figure*}[!hbt]
    \begin{center}
        \subfloat[Ant-v0 (PyBullet)]{\label{fig-ant-eq9}\includegraphics[width=0.33\linewidth]{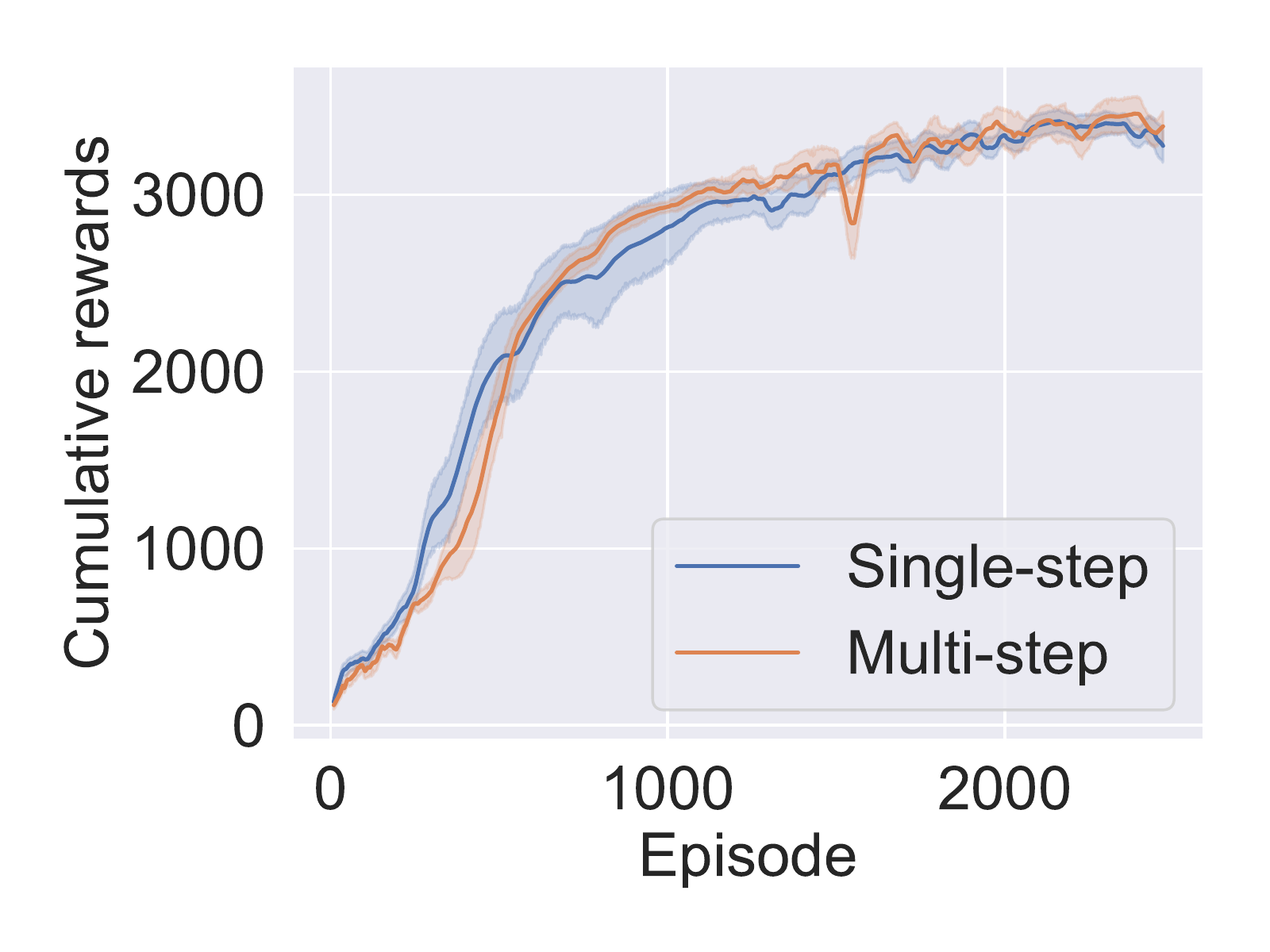}}
        \subfloat[Hopper-v0 (PyBullet)]{\label{fig-hc-eq9}\includegraphics[width=0.33\linewidth]{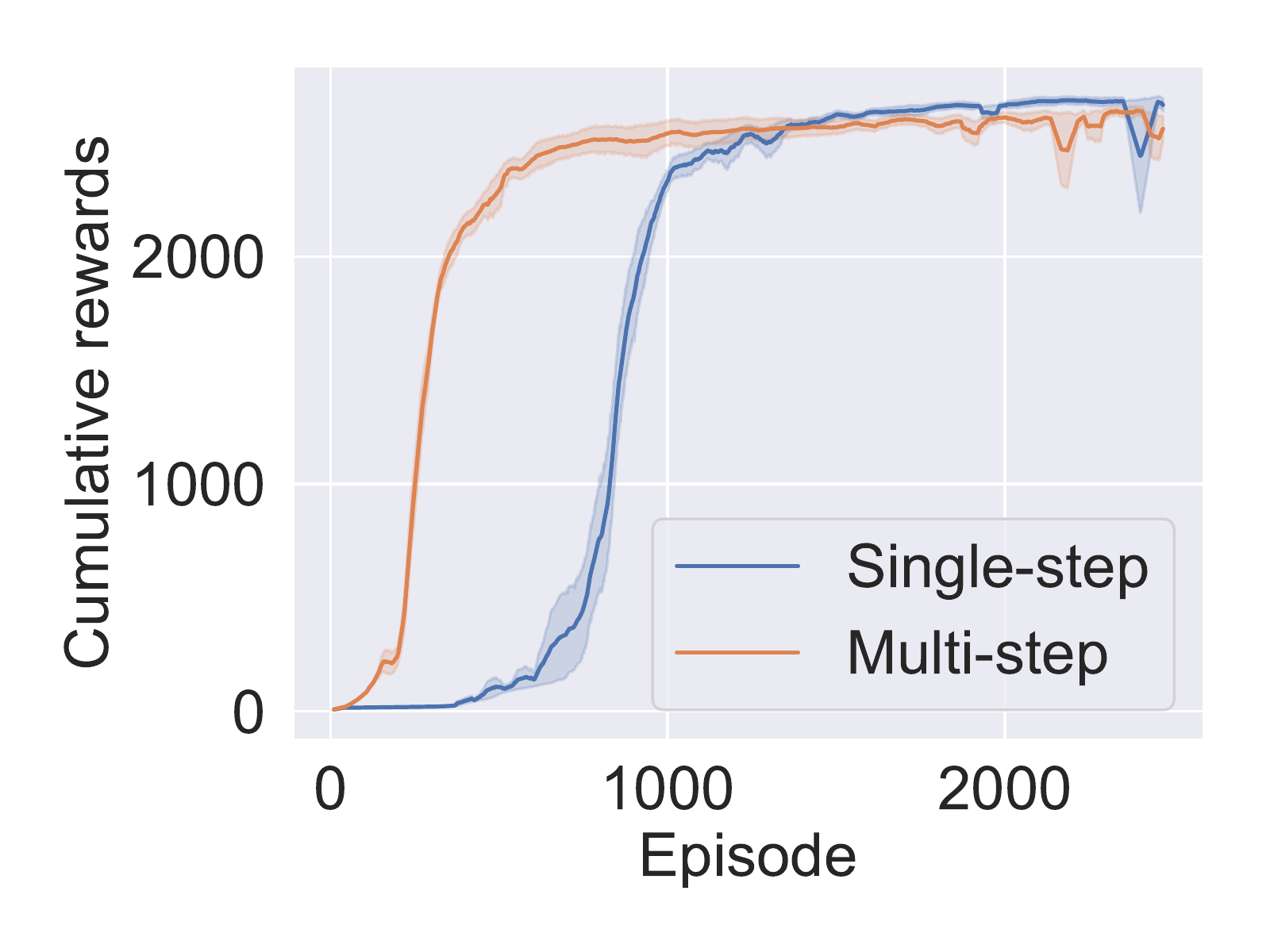}}
      \subfloat[Walker2D-v0 (PyBullet)]{\label{fig-reacher-eq9}\includegraphics[width=0.33\linewidth]{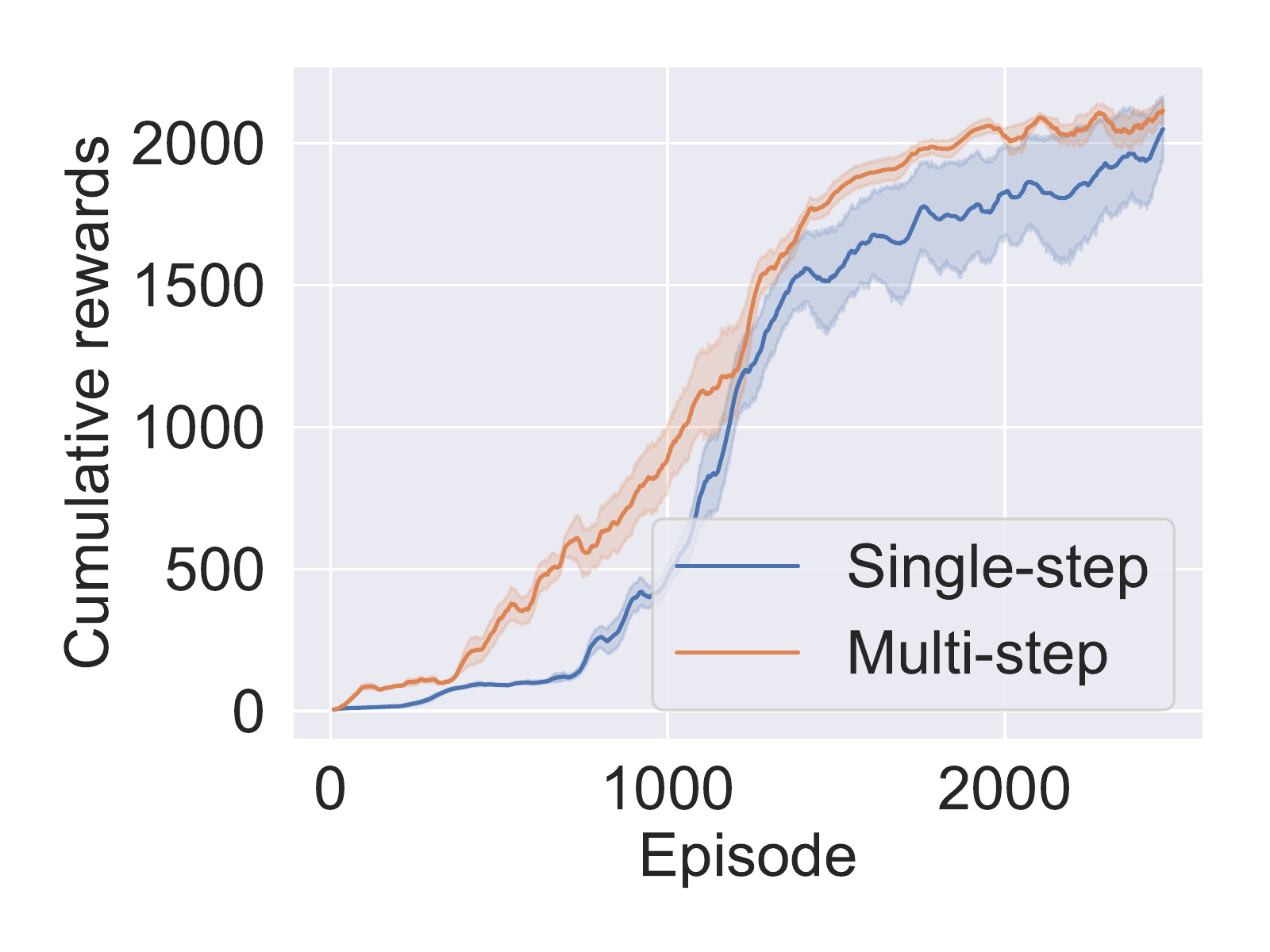}}\\
       \subfloat[Hopper-v3 (Mujoco)]{\label{fig-hopper-eq9}\includegraphics[width=0.33\linewidth]{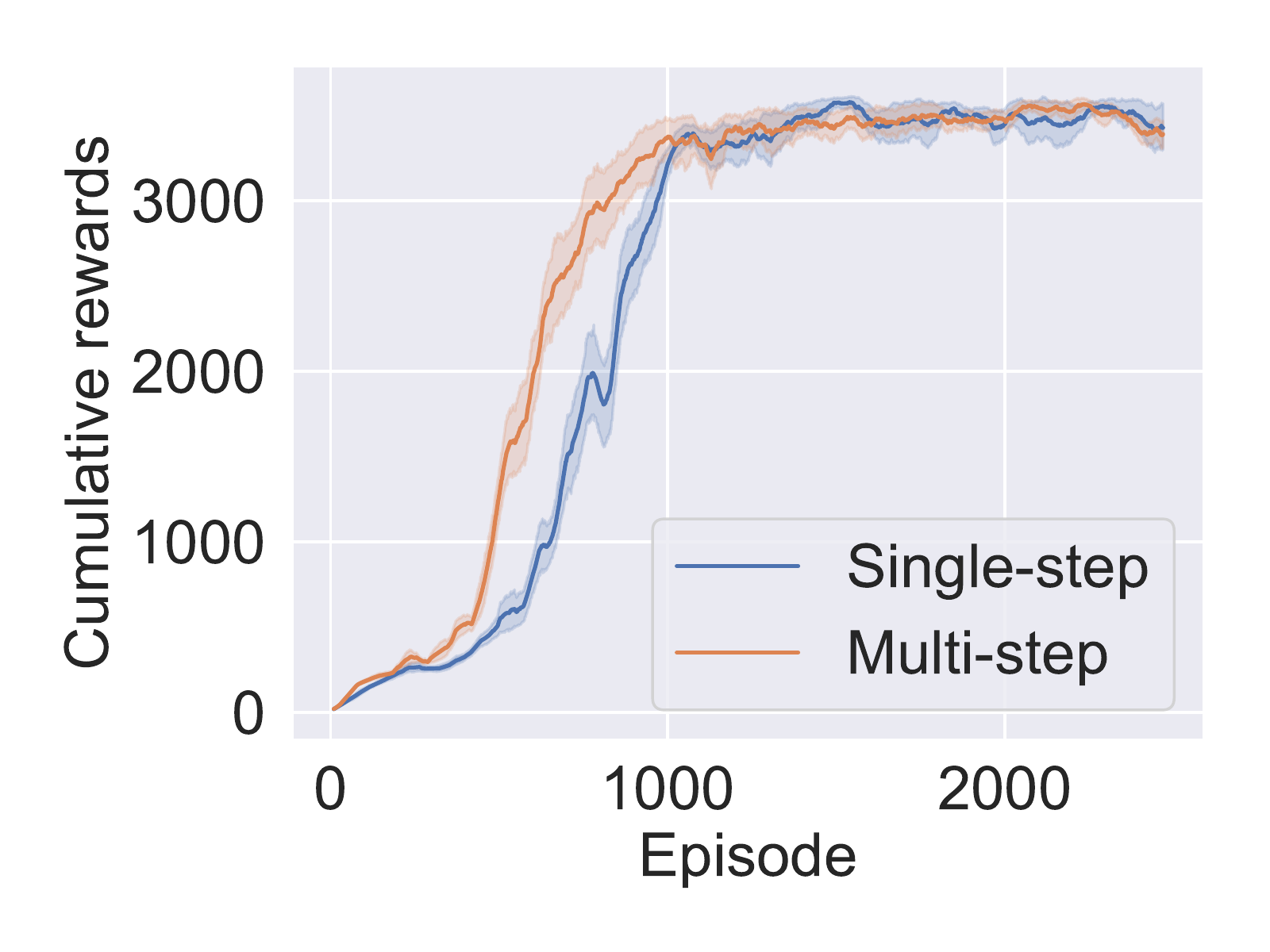}}
      \subfloat[InvertedDoublePendulum-v2 (Mujoco)]{\label{fig-idpPB-eq9}\includegraphics[width=0.33\linewidth]{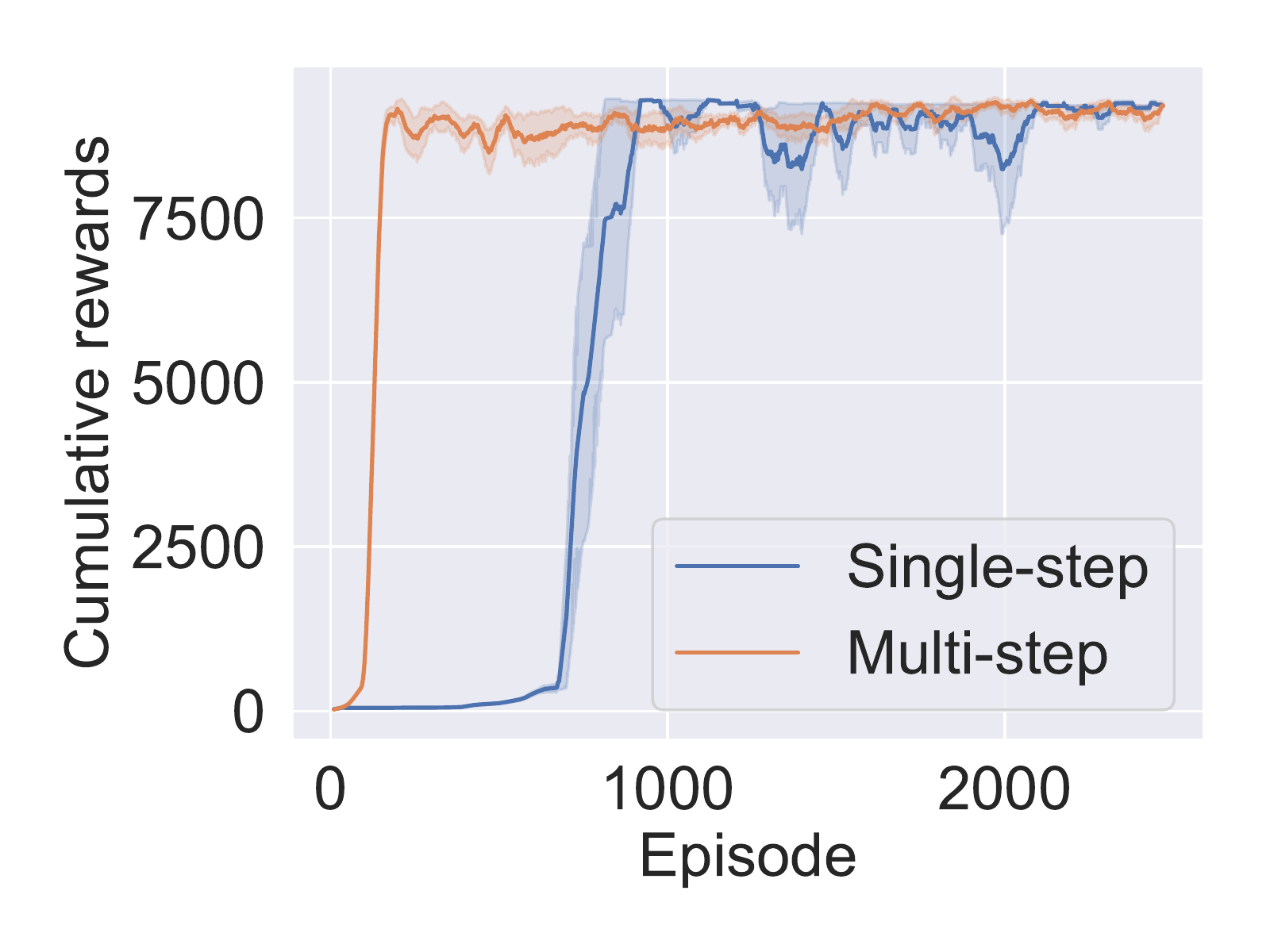}}
       \subfloat[Walker2D-v3 (Mujoco)]{\label{fig-walker-eq9}\includegraphics[width=0.33\linewidth]{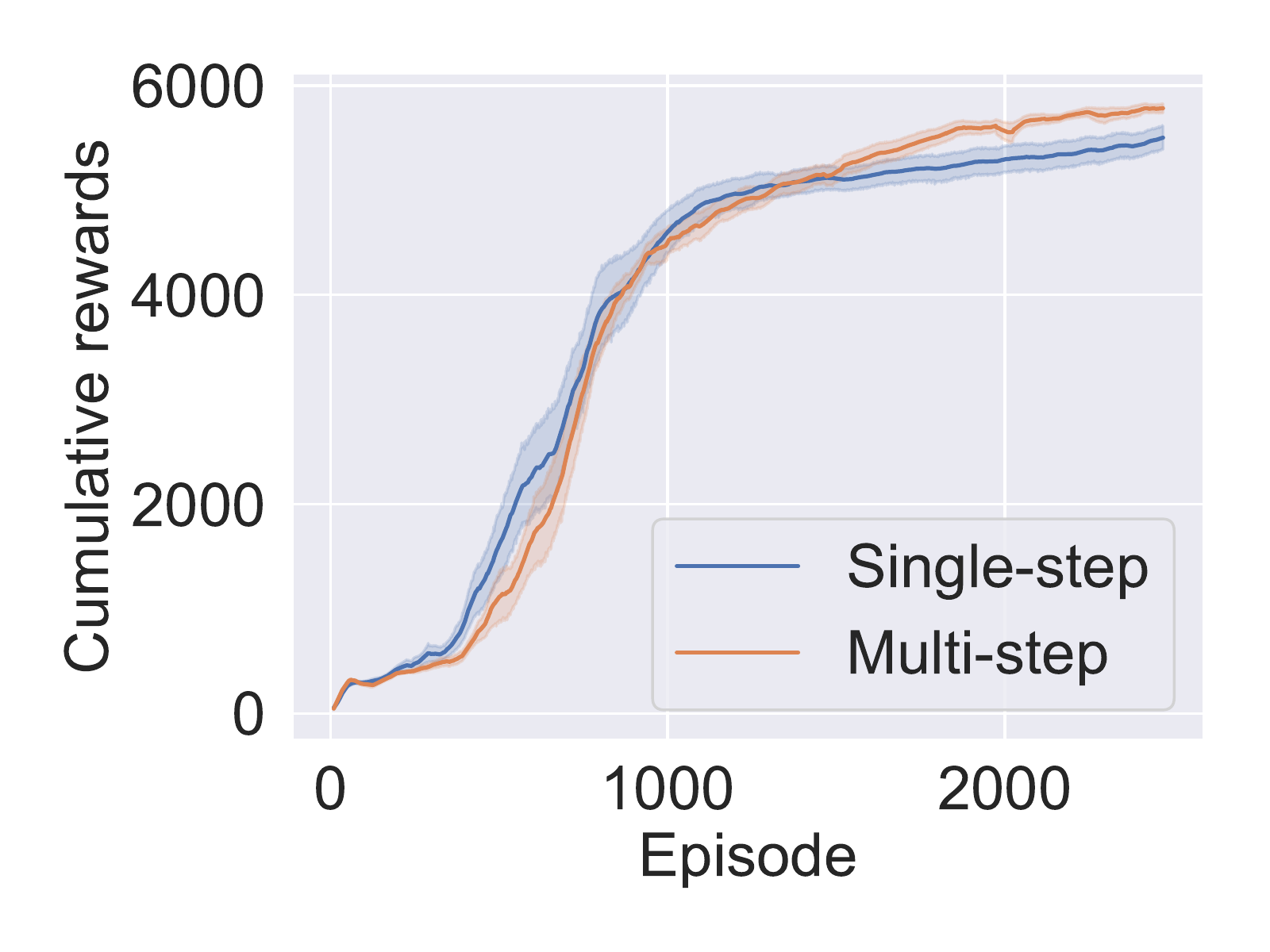}}
       \\
\end{center}
    \caption{Learning curves of HED using the single-step high-level policy training technique in \eqref{equ-e-pg} vs. the proposed multi-step high-level policy training technique in \eqref{equ-mu-new}.}
    \label{fig:eq9_impact}
\end{figure*}

All experiments were run using a cluster of Linux computing nodes. Each node is equipped with 16 GB memory. The CPU specification is provided in Table~\ref{tab:cpu-para}. Each experiment was run in a Python virtual environment managed by Anaconda with Python packages specified in Table~\ref{tab:python-lib}. Based on the above hardware and software configurations, we found that the running time required by HED on all experimented benchmarks ranges from 2 hours to 22 hours. Other ensemble learning algorithms, including both ED2 and SUNRISE, require similar running time on each benchmark. This observation indicates that HED does not noticeably increase the algorithm running time, compared to other previously proposed ensemble algorithms.

\begin{table}[htb!]
\caption{CPU specification.}
\label{tab:cpu-para}
\centering
\begin{tabular}{ll}
\hline
Architecture       & x86\_64                                        \\
CPU op-mode(s)     & 32-bit, 64-bit                                 \\
CPU(s)             & 16                                             \\
CPU family         & 6                                              \\
Thread(s) per core & 2                                              \\
CPU max MHz        & 4900.0000                                      \\
CPU min MHz        & 800.0000   \\
Model name         & 11th Gen Intel(R) Core(TM)\\ 
 & i7-11700 @ 2.50GHz  \\
\hline
\end{tabular}
\end{table}

\begin{table}[htb!]
\caption{Python packages.}
\label{tab:python-lib}
\centering
\begin{tabular}{ll}
\hline
Package name      &Version    \\ \hline
cython    &0.29.25    \\
gym       &0.21.0     \\
keras     &2.7.0     \\
mujoco-py   &2.1.2.14    \\
numpy    &1.21.4   \\
pybulletgym       &0.1 \\
python    &3.7.11       \\
scipy      &1.7.3      \\
tensorflow    &2.7.0 \\
\hline
\end{tabular}
\end{table}

\section{Performance Impact of High-Level Policy Training Frequencies}

In this appendix, we study the effectiveness of conducting high-level policy training after HED obtains a full sampled episode. Figure~\ref{fig:when2HLtrain_impact} shows the performance comparison of HED with two different training frequencies: every 50 consecutive timesteps vs. every episode. It can be noticed that, performing high-level policy training after every episode (orange curve) can significantly improve the HED algorithm in terms of both the final performance and convergence speed. For example, as shown in Figure~\ref{fig:when2HLtrain_impact}(c), the orange curve reaches 2000 before 2000 episodes while the blue curve stays below 1500 and fluctuates between 1000 and 1500 after 2000 episodes.

We also notice that episodic policy training is more robust to the randomness in the environment and less sensitive to the initialization of neural network weights. For example, in Figure~\ref{fig:when2HLtrain_impact}(d), episodic policy training produces a smaller confident interval (orange shaded area) compared to the fixed timestep training (blue shaded area) over 10 independent algorithm runs. Similar results can also be observed from Figures~\ref{fig:when2HLtrain_impact}(a), \ref{fig:when2HLtrain_impact}(b), and \ref{fig:when2HLtrain_impact}(e). Note that in each algorithm run, both policy networks and Q-networks are initialized with different weights. The environment initial states also vary.

\section{Effectiveness of Multi-Step High-Level Policy Training}

This appendix investigates the effectiveness of multi-step policy training by using \eqref{equ-mu-new}. Specifically, we compare the performance of HED against its variant, which performs single-step high-level policy training by using \eqref{equ-e-pg}, on 6 problems that include both PyBullet and Mujoco benchmarks. 



As shown in Figure~\ref{fig:eq9_impact}, the proposed multi-step policy training technique converges clearly faster and is more stable during the learning process than the single-step training technique. In Figure~\ref{fig:eq9_impact}(b), the orange curve converges after 500 episodes while the blue curve converges after 1000 episodes. The significant improvement in convergence speed can also be witness in Figure~\ref{fig:eq9_impact}(e).

The shaded areas in Figures~\ref{fig:eq9_impact}(a), \ref{fig:eq9_impact}(c), and \ref{fig:eq9_impact}(e) also show that the multi-step training technique is less sensitive to the environment randomness and neural network weight initialization, compared to using the conventional single-step training method in \eqref{equ-e-pg}. Hence, our experiment results demonstrate the importance of inter-learner collaboration. By enabling base learners in an ensemble to explicitly share their learned policy parameters, HED can achieve high convergence speed and effectively boost the learning process.

\section{Additional Experiment Results}
This appendix presents additional experiment results. Specifically, Table~\ref{tab:perf_comp_DPD-PPO_vs_HED} compares the performance of DPD-PPO reported in~\cite{lai2020dual} directly with the results of HED. It shows that HED can outperform DPD-PPO, even when DPD-PPO used four times as many state-transition samples as HED.
\begin{table}[!ht]
\caption{Comparison between HED and the performance of DPD-PPO reported in \protect\cite{lai2020dual} on two commonly tested benchmarks.}
\centering
\resizebox{.4\textwidth}{!}{
\begin{tabular}{c||cc}
\hline
\emph{Mean}                       & DPD-PPO                    & HED             \\ \hline 
Humanoid            & 2242.19 \cite{lai2020dual} &    \textbf{4223.77}  \\
Walker2D             &  3857.23 \cite{lai2020dual}  &   \textbf{5778.84}  \\ \hline
\emph{Max}                       & DPD-PPO                     & HED             \\ \hline 
Humanoid  & 3885.83 \cite{lai2020dual}  & \textbf{5553.51} \\
Walker2D  & 5233.56 \cite{lai2020dual}  & \textbf{5991.81}\\ \hline
\end{tabular}}
\label{tab:perf_comp_DPD-PPO_vs_HED}
\end{table}

Following existing works \cite{lai2020dual}, besides the average cumulative reward, we also compare the maximum cumulative rewards achieved by all competing algorithms over 10 independent runs on 9 benchmark problems. Table~\ref{tab:final_perf_comp} shows that HED achieves the highest maximum cumulative rewards on 7 out of 9 benchmark problems. For the remaining two problems (InvertedDoublePendulum-v2 and Walker2D-v3), HED proves to be a strong contender by achieving the second highest rewards among all competing algorithms.

\begin{table}[!th]
\caption{Maximum cumulative rewards over 10 independent algorithm runs of all competing algorithms on 9 benchmark problems.}
\vspace{-8pt}
\centering
\resizebox{.49\textwidth}{!}{
\begin{tabular}{c||cccc|c}
\hline
Benchmark problems                       & TD3            & SAC            & ED2             & SUNRISE         & HED             \\ \hline 
Ant-v0 (PyBullet)              & 3595  & 3172  &  3630  &  3577  &  \textbf{3645}\\
Hopper-v0 (PyBullet)              & 2752 & 2342  & 2632  & 2593  & \textbf{2754}  \\
InvertedPendulum-v0 (PyBullet)       & \textbf{1000}   & \textbf{1000}   &  \textbf{1000}    & \textbf{1000}   & \textbf{1000}      \\
Walker2D-v0 (PyBullet)              & 2038  & 957    &  1982  &  2234 & \textbf{2294}\\
Hopper-v3 (Mujoco)             & 3359 & 3600 & 3563  & 3281  & \textbf{3647}   \\
Humanoid-v3   (Mujoco)            & 748  &  2837 & 840 &  1059  & \textbf{5553}  \\
InvertedDoublePendulum-v2 (Mujoco)               & 9307    & \textbf{9359}   & 9350   & \textbf{9359}   & 9357   \\
LunarLanderContinuous-v2         & 282  &  282 &  284  &  286  & \textbf{287} \\
Walker2D-v3 (Mujoco)             & 5407 & 5275  &  5335 & \textbf{6716} &  5991  \\ \hline
\end{tabular}}
\label{tab:final_perf_comp}
\vspace{-10pt}
\end{table}

\end{document}